\documentclass[runningheads]{llncs}
\usepackage[]{appendix}

\usepackage{graphicx}
\usepackage{caption}
\usepackage{subfig}
\usepackage{amsmath,amssymb} 
\usepackage{color}
\usepackage[width=122mm,left=12mm,paperwidth=146mm,height=193mm,top=12mm,paperheight=217mm]{geometry}
\usepackage{times}
\usepackage{epsfig}
\usepackage{graphicx}
\usepackage{amsmath}
\usepackage{amssymb}
\usepackage{booktabs}       
\usepackage{standalone}
\usepackage{xspace}
\usepackage{multirow}
\usepackage{diagbox}
\usepackage{wrapfig}
\newcommand{\fact}[1]{$\langle$#1$\rangle$}
\def \hfillx {\hspace*{-\textwidth} \hfill}

 \makeatletter
 \DeclareRobustCommand\onedot{\futurelet\@let@token\@onedot}
 \def\@onedot{\ifx\@let@token.\else.\null\fi\xspace}
 \def\eg{e.g\onedot} 
 \def\ie{i.e\onedot}

 \makeatother
 \newcommand{\myparagraph}[1]{\noindent\textbf{#1}}
\usepackage[dvipsnames]{xcolor}

\begin{document}
\title{Memory Aware Synapses: Learning what (not) to forget} 

\titlerunning{Memory Aware Synapses: Learning what (not) to forget}
%
\author{Rahaf Aljundi\inst{1}\and
  Francesca Babiloni                   \inst{1} \and
Mohamed Elhoseiny\inst{2} \and \\
Marcus Rohrbach\inst{2} \and
Tinne Tuytelaars\inst{1} }
\authorrunning{R. Aljundi, F. Babiloni, M. Elhoseiny, M. Rohrbach and T. Tuytelaars}
%

\institute{KU Leuven, ESAT-PSI, iMEC, Belgium 
\and Facebook AI Research}
\maketitle              
\begin{abstract}
Humans can learn in a continuous manner. Old rarely utilized  knowledge can be overwritten by new incoming information while important, frequently used knowledge is prevented from being erased. In artificial learning systems, lifelong learning so far has focused mainly on accumulating knowledge over tasks and overcoming catastrophic forgetting. In this paper, we argue that, given the limited model capacity and the unlimited 
new information to be learned, knowledge has to be preserved or erased selectively. 
Inspired by neuroplasticity, we propose a novel approach for lifelong learning, coined Memory Aware Synapses ({\tt MAS}). It computes the importance of the parameters of a neural network in an unsupervised and online manner. Given a new sample which is fed to the network, {\tt MAS} accumulates an importance measure for each parameter of the network, based on how sensitive the predicted output function is to a change in this parameter.
When learning a new task, changes to important parameters can then be penalized, effectively preventing important knowledge related to previous tasks from being overwritten.  Further, we show an interesting connection between a local version of our method and
Hebb's rule,
which is a model 
for the learning process in the  brain. We test our method on a sequence of object recognition tasks and on the challenging problem of learning an embedding
for predicting $<$subject, predicate, object$>$ triplets. We show state-of-the-art performance and, for the first time, the ability to adapt the importance of the parameters based on unlabeled data towards what the network needs (not) to forget, which may vary depending on  test conditions. 

\end{abstract}

\section{Introduction}\label{introduction}
The real and digital world around us evolves continuously. Each day millions of images with new tags appear on social media. Every minute hundreds of hours of video are uploaded on Youtube. This new content contains new topics and trends that may be very different from what one has seen before - think  e.g.~of new emerging news topics, fashion trends, social media hypes or technical evolutions.
Consequently, to keep up to speed, { our learning systems should be able to evolve as well}. 

Yet the dominating paradigm to date, using supervised learning, ignores this issue. 
It learns a given task using an existing set of training examples. Once the training is finished, the trained model is frozen and deployed. 
From then on, new incoming data is processed without any 
further adaptation or customization of the model. Soon, the model becomes outdated. In that case, the training process has to be repeated, using both the previous and new data, and with an extended set of category labels.
In a world like ours, such a practice becomes 
intractable when moving to real scenarios such as those mentioned earlier, where the data is streaming, might be disappearing after a given period of time or even can't be stored at all due to storage constraints or privacy issues. 

In this setting, lifelong learning (LLL)~\cite{pentina15nips,silver2013lifelong,thrun1995lifelong} comes as a natural solution. LLL studies continual learning across tasks and data, tackling one task at a time, without storing data from previous tasks. The goal is to accumulate knowledge across tasks (typically via model sharing), resulting in a single model that  performs well on all the learned tasks. The question then is how to overcome catastrophic forgetting~\cite{french1999catastrophic,goodfellow2013empirical,mccloskey1989catastrophic} of the old knowledge when starting a new learning process using the same model.

\begin{figure*}[t]
\centering
\includegraphics[width=0.95\textwidth]{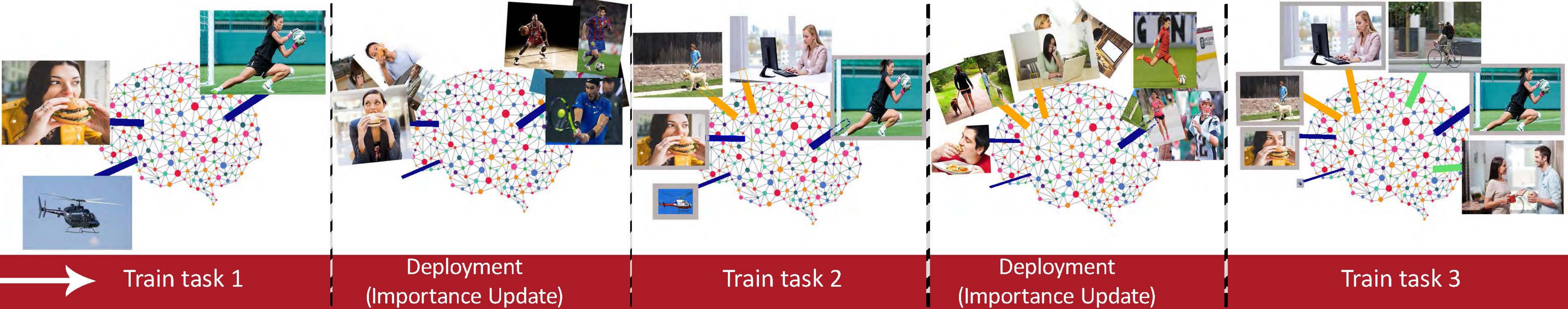} ...
  \caption{
Our continuous learning setup. As common in the LLL literature, tasks are learned in sequence, one after the other. 
If, in between learning tasks, the agent is active and performs the learned tasks, we can use these unlabeled samples 
to update importance weights for the model parameters. 
Data that appears frequently, will have a bigger contribution. This way, the agent learns 
what is
important and should not be forgotten. 
}  
\vspace*{-20pt}
  \label{fig:teaser}
\end{figure*}

So far, LLL methods have mostly (albeit not exclusively) 
been applied to relatively short sequences -- often consisting of no more than two tasks (\eg~\cite{lee2017overcoming,li2016learning,rannen2017encoder}), and using relatively large networks with plenty of capacity (\eg~\cite{aljundi2016expert,fernando2017pathnet,rusu2016progressive}). 
However, in a true LLL setting with a never-ending list of tasks, the capacity of the model sooner or later reaches its limits and compromises need to be made. 
Instead of aiming for no forgetting at all, figuring out what can possibly be forgotten becomes at least as important.
In particular, exploiting context-specific test conditions may pay off in this case. Consider for instance a surveillance camera. Depending on how or where it is mounted, it always captures images under particular viewing conditions. Knowing how to cope with other conditions is no longer relevant and can be forgotten, freeing capacity for other tasks. This calls for a LLL method that can learn what (not) to forget using unlabeled  test data. We illustrate this setup in Figure~\ref{fig:teaser}.

Such adaptation and memory organization is what we also observe in  biological neurosystems. Our ability to preserve what we have learned before is largely dependent on how frequent we make use of it. Skills that we practice often, appear to be unforgettable, unlike those that we have not used for a long time.
Remarkably, this flexibility and adaptation occur
in the absence of any form of supervision.
According to Hebbian theory~\cite{hebb2002organization}, the process at the basis of this phenomenon is the strengthening of synapses connecting neurons that fire synchronously, compared to those connecting neurons with unrelated firing behavior.

In this work, we propose a new method for LLL, coined {\em Memory Aware Synapses}, or {\tt MAS} for short, inspired by the model of Hebbian learning in biological systems. Unlike previous works, {\em our LLL method can learn 
what parts of the model are important 
using unlabelled data}. This allows for  
adaptation to specific test conditions and continuous updating of importance weights. 
This is achieved by estimating  importance weights for the network parameters without relying on the loss, but by looking at the sensitivity of the output function instead. This way, our method not only avoids the need for labeled data, but importantly it also avoids complications due to the loss being in a local minimum, resulting in gradients being close to zero.
This makes our method not only more versatile, but also simpler, more memory-efficient, and, as it turns out, more effective in learning what not to forget, compared to other model-based LLL approaches.

\textbf{Contributions} of this paper are threefold:\ 
{\em First}, we propose a new LLL method {\em Memory Aware Synapses} (MAS). It estimates importance weights for all the network parameters in an unsupervised and online manner,
allowing adaptation to unlabeled data, \eg in the actual test environment. \ {\em Second}, we show how a local variant of MAS 
is linked to the Hebbian learning scheme.\ 
{\em Third}, we achieve better performance than state-of-the-art, both when using the standard LLL setup and when adapting to specific test conditions, both for object recognition and for predicting $<$subject, predicate, object$>$ triplets, where an embedding is used instead of a softmax output.

In the following we discuss related work 
in section \ref{sec:related_work} and give some background information
in section~\ref{sec:background}. Section \ref{sec:method} describes our method  and its connection with Hebbian learning.
Experimental results are given in section \ref{sec:experiments} and section \ref{sec:conculosion} concludes the paper.

\begin{wraptable}{r}{0.54\textwidth}
\vspace{-1.4cm}
  \begin{scriptsize}
   \begin{tabular}{|l|c|c|c|c|c|c|}
   \hline
Method & Type & Constant & Problem & On   Pre-       & Unlabeled  & Adap-\\
       &      & Memory   & agnostic & trained & data & tive\\\hline
       \texttt{LwF} \cite{li2016learning} & data  & \checkmark & \text{X} & \checkmark & n/a & \text{X}\\
       \texttt{EBLL} \cite{rannen2017encoder} & data& \text{X} & \text{X} & \text{X} & n/a & \text{X}\\
        \texttt{EWC} \cite{kirkpatrick2016overcoming} & model &\checkmark& \checkmark & \checkmark &\text{X} & \text{X}\\
        \texttt{IMM} \cite{lee2017overcoming} & model  & \text{X} & \checkmark & \checkmark &\text{X} & \text{X}\\
        \texttt{SI} \cite{zenke2017improved} & model  & \checkmark &  \checkmark &\text{X}&\text{X} & \text{X}\\
        \texttt{MAS} (our) & model  & \checkmark &\checkmark& \checkmark &\checkmark & \checkmark\\ \hline
\end{tabular}
\end{scriptsize}
\vspace{-0.2cm}
  \caption{\footnotesize LLL desired characteristics and the compliance of methods, that treat forgetting without storing the data, to these characteristics. }
    \label{tbl:checklist}
    \vspace{-0.4cm}
\end{wraptable}

\section{Related Work}\label{sec:related_work}
While lifelong learning has been studied since a long time in different domains (\eg robotics~\cite{thrun1995lifelong} or machine learning~\cite{ring1997child}) and touches upon the broader fields of meta-learning~\cite{finn17icml}  and learning-to-learn \cite{andrychowicz16nips}, we focus in this section on more recent work in the context of computer vision mainly. 

The  main challenge in LLL is to adapt the learned model continually to new tasks, be it from a similar or a different environment~\cite{pentina2015lifelong}. 
However, looking at existing LLL solutions, we observe that none of them satisfies all the characteristics one would expect or desire from a lifelong learning approach (see Table~\ref{tbl:checklist}). 
First, its memory should be constant w.r.t. the number of tasks, to avoid a gradual increase in memory consumption over time.
not be limited to a specific setting (e.g. only classification). We refer to this as problem agnostic. 
Third, given a pretrained model, it should be able to build on top of it and add new tasks. 
Fourth, being able to learn from unlabeled data 
would increase the method applicability to cases where original training data no longer exists.
Finally, as argued above, within a fixed capacity network, being able to adapt what not to forget to a specific user setting 
would leave more free capacity for future tasks.
In light of these properties, we discuss recently proposed methods. They can be divided into  two main approaches:
data-based and model-based approaches. Here, we don't consider LLL methods that require storing samples, such as~\cite{lopez2017gradient,rebuffi2016icarl}.

\noindent
{\bf Data-based approaches}~\cite{aljundi2016expert,li2016learning,rannen2017encoder,shmelkov17iccv}
use data from the new task to approximate the performance of the previous tasks. This works best if the data distribution mismatch between tasks is limited.
Data based approaches are mainly designed for a classification scenario and overall, the need of these approaches to have a preprocessing step before each new task, to record the targets for the previous tasks is an additional limitation.

\noindent{\bf Model-based approaches}~\cite{fernando2017pathnet,kirkpatrick2016overcoming,lee2017overcoming,zenke2017improved}, like our method, focus on the parameters of the network instead of depending on the task data.
Most similar to our work are \cite{kirkpatrick2016overcoming,zenke2017improved}. Like them, we estimate an importance weight for each model parameter and add a regularizer when training a new task that penalizes any changes to important parameters. The difference lies in the way the importance weights are computed. In the \textit{Elastic Weight Consolidation} work~\cite{kirkpatrick2016overcoming}, this is done based on an approximation of the diagonal of the Fisher information matrix. 
In the \textit{Synaptic Intelligence} work~\cite{zenke2017improved}, importance weights are computed during training in an online manner. To this end, they record how much the loss would change due to a change in a specific parameter and accumulate this information over the training trajectory. 
However,  also this method has some drawbacks: 1) Relying on the weight changes in a batch gradient descent might overestimate the importance of the weights, as noted by the authors. %
2) When starting from a pretrained network, as in most practical computer vision applications, some weights might be used without big changes. As a result, their importance will be underestimated. %
3) The computation of the importance is done during training and fixed later.  In contrast, we believe the importance of the weights should be able to adapt to the test data where the system is applied to. 
In contrast to the above two methods, we propose to look at the sensitivity of the learned function, rather than the loss. This simplifies the setup considerably since, unlike the loss, the learned function is not in a local minimum, so complications with gradients being close to zero are avoided. 

In this work, we propose a model-based method that computes the importance of the network parameters not only in an online manner but also adaptive to the data that the network is tested on  in an unsupervised manner. While previous works \cite{quadrianto2009distribution,royer2015classifier} adapt the learning system  at prediction time in a transductive setting, our goal here is to build a continual system that can adapt the importance of the weights to what the system needs to remember.  
Our method requires a constant amount of memory and enjoys the main desired characteristics of lifelong learning we listed above while achieving state-of-the-art performance.

\section{Background}\label{sec:background}
\noindent
{\bf Standard LLL setup.}
Before introducing our method, we briefly remind the reader of the standard LLL setup, as used, e.g., in~\cite{aljundi2016expert,lee2017overcoming,li2016learning,rannen2017encoder,zenke2017improved}.
It  focuses on image classification and consists of a sequence of disjoint {\em tasks}, which are learned one after the other. Tasks may correspond to different datasets, or different splits of a dataset, without overlap in category labels. %
The assumption of this setup is that, when training a task, 
only the data related to that task is accessible.  
Ideally, newer tasks can benefit from the representations learned by older tasks (forward transfer). Yet in practice, the biggest challenge is to avoid catastrophic forgetting of the old tasks' knowledge (i.e., forgetting how to perform the old tasks well). This is a far more challenging setup than joint learning, as typically used in the multitask learning literature, where all tasks are trained simultaneously. 

\noindent
{\bf Notations.} 
We train a single, shared neural network over a sequence of tasks. The parameters $\{\theta_{ij}\}$ of the model are the weights of the connections between pairs of neurons $n_i$ and $n_j$ in two consecutive layers\footnote{In convolutional layers, parameters are shared by multiple pairs of neurons. For the sake of clarity, yet without loss of generality, we focus here on fully connected layers.}. %
As in other model-based approaches, our goal is then to compute an importance value $\Omega_{ij}$ 
for each parameter $\theta_{ij}$,
indicating its importance with respect to the previous tasks. 
In a learning sequence, we receive a sequence of tasks $\{T_n\}$ to be learned, each with its training data ($X_n,\hat{Y}_n$), with $X_n$ the input data and $\hat{Y}_n$
the corresponding ground truth output data (labels). 
Each task comes with a task-specific loss $L_n$, that will be combined with an extra loss term to avoid forgetting.
When the training procedure converges to a local minimum, the model has learned an approximation $F$ of the true function $\bar{F}$. 
$F$  maps a new input $X$ to the outputs ${Y_1, ... ,Y_n}$ for tasks $T_1 ... T_n$ learned so far.

\section{Our Approach}
\label{sec:method}
\begin{figure}[t]
\centering
\includegraphics[width=0.8\linewidth]{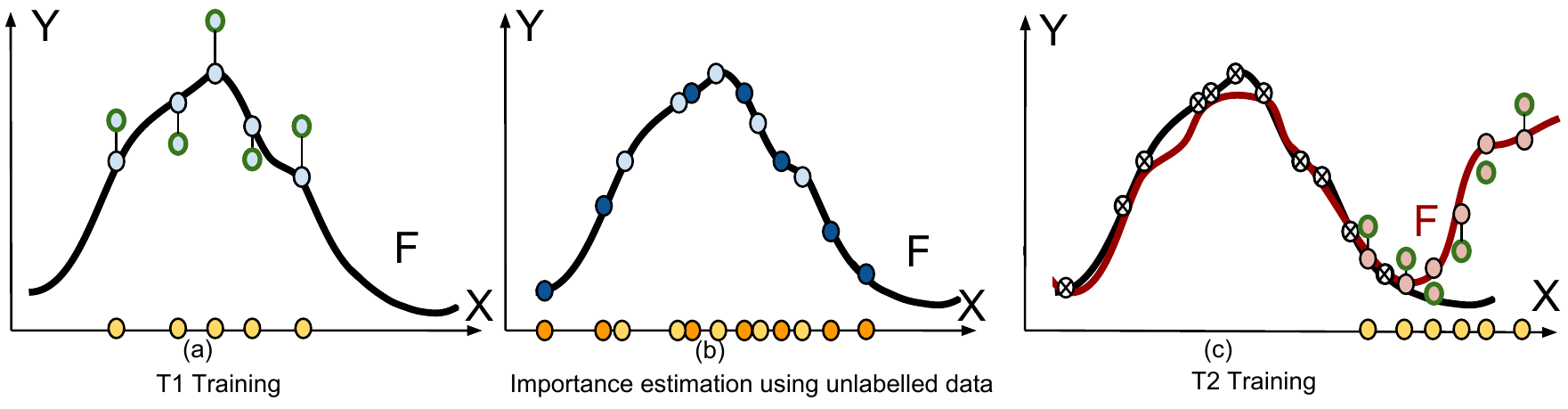}
  \caption{\footnotesize \cite{zenke2017improved,kirkpatrick2016overcoming} estimate the parameters importance based on the loss, comparing the network output (light blue) with the ground truth labels (green) using training data (in yellow) (a). In contrast, we estimate the parameters importance, after convergence, based on the sensitivity of the learned function to their changes (b). This allows using additional unlabeled data points (in orange). When learning a new task, changes to important parameters are penalized,  the function is preserved over the domain densely sampled in (b), while adjusting not important parameters to ensure good performance on the new task (c).} 
  \label{fig:Function_preserve}
 \vspace*{-10pt} 
\end{figure}

In the following, we introduce our approach.
Like other model-based approaches~\cite{kirkpatrick2016overcoming,zenke2017improved}, we estimate an importance weight for each parameter in the network.
Yet in our case, these importance weights approximate the {\em sensitivity of the learned function} to a parameter change  rather than a measure of the (inverse of) parameter uncertainty, as in~\cite{kirkpatrick2016overcoming}, or the sensitivity of the loss to a parameter change, as in~\cite{zenke2017improved} (see Figure~\ref{fig:Function_preserve}). As it does not depend on the ground truth labels, our approach allows computing the importance using any available data (unlabeled) which in  turn allows for an adaptation to user-specific settings.
In a learning sequence, we start with task $T_1$, training the model to minimize the task loss $L_1$ on the training data ($X_1$, $\hat{Y_1}$) -- or simply using a pretrained model for that task.
\subsection{ Estimating parameter importance}
After convergence,  the model has learned an approximation $F$ of the true function $\bar{F}$. $F$  maps the input $X_1$ to the output $Y_1$.  This mapping $F$ is the target we want to preserve while learning additional tasks. %
To this end, we measure how sensitive the  function $F$ output is to changes in the network parameters. 
For a given data point 
$x_k$,
the output of the network is $F(x_k; \theta)$. A small perturbation $\delta = \{\delta_{ij}\}$ in the parameters $\theta = \{\theta_{ij}\}$ results  in a change in the function output that can be approximated by: 
\begin{equation}\label{eq:1}
   F(x_k; \theta+ \delta) -F(x_k; \theta)\approx \sum_{i,j} g_{ij}(x_k)  \delta_{ij}
\end{equation}
where $g_{ij}(x_k)=\frac{\partial (F(x_k; \theta))}{\partial \theta_{ij}}$ is the gradient of the learned function  with respect to the parameter $ \theta_{ij}$  evaluated at the data point $x_k$ and $\delta_{ij}$ is the change in parameter $\theta_{ij}$.
Our goal is to preserve the prediction of the network (the learned function) 
at each observed data point and prevent changes to parameters that are important for this prediction.

Based on equation~\ref{eq:1} and assuming a  small constant change $\delta_{ij}$, we can measure the importance of a parameter by the magnitude of the gradient $g_{ij}$, i.e.~how much  does a small perturbation to that parameter  change the output of the learned function for data point $x_k$. 
We then accumulate the gradients over the given data points %
to obtain importance weight $\Omega_{ij}$ for parameter $\theta_{ij}$: 
\begin{equation}\label{eq:2}
\Omega_{ij} =\frac{1}{N}\sum_{k=1}^N  \mid\mid g_{ij}(x_k) \mid\mid
\end{equation}
This equation can be updated in an online fashion whenever a new data point is fed to the network. $N$ is the total number of data points at a given phase. 
Parameters with small importance weights do not affect the output much, and can, therefore, be changed to minimize the loss for subsequent tasks, while parameters with large weights should ideally be left unchanged.

When the output function $F$ is multi-dimensional, as is the case for most neural networks, equation~\ref{eq:2} involves computing the gradients for each output, which requires as many backward passes as the dimensionality of the output. As a more efficient alternative, we propose to use the gradients of the squared $\ell_2$ norm of the learned function output\footnote{We   square the $\ell_2$ norm as it simplifies the math and the link with the Hebbian method, see section~\ref{subsec:hebb}. }, \ie, $g_{ij}(x_k)=\frac{\partial [\ell^2_2(F(x_k; \theta))]}{\partial \theta_{ij}}$.
The importance of the parameters is then measured by the sensitivity of the squared $\ell_2$ norm of the function output to their changes. This way, we get one scalar value for each sample instead of a vector output. Hence, we only need to compute one backward pass and can use the resulting gradients for estimating the parameters importance.
Using our method,  for regions in the input space that are sampled densely,
the function will be preserved and catastrophic forgetting is avoided. However, parameters not affecting those regions will be given low importance weights, and can be used to optimize the function for other tasks, affecting the function over other regions of the input space.  

\subsection{ Learning a new task}
When a new task $T_n$ needs to be learned, we have in addition to the new task loss $L_n(\theta)$, a regularizer that penalizes changes to parameters that are deemed important for previous tasks:

\begin{equation}\label{eq:3}
L(\theta)=L_n(\theta)+{\lambda} \sum_{i,j} \Omega_{ij}(\theta_{ij}-\theta^*_{ij})^2
\end{equation}
with $\lambda$ a hyperparameter for the regularizer and $\theta_{ij}^*$ the ``old'' network parameters (as determined by the optimization for the previous task in the sequence, $T_{n-1}$).  As such we allow the new task to change parameters that are not important for the previous task (low $\Omega_{ij}$).  The important parameters (high $\Omega_{ij}$) can also be reused, via model sharing, but with a penalty when changing them.

Finally, the importance matrix $\Omega$  is to be updated after training a new task, by accumulating over the previously computed $\Omega$. Since we don't  use the loss function, $\Omega$ can be computed on any available data considered most representative for test conditions, be it on the last training epoch, during the validation phase or at test time. 
In the experimental section~\ref{sec:experiments}, we show how this allows our method to adapt and specialize to any set, be it from the training or from the test.

\subsection{Connection to Hebbian learning}\label{subsec:hebb}
\begin{wrapfigure}{r}{0.4\textwidth}
    \vspace*{-35pt}
    \includegraphics[width=0.4\textwidth]{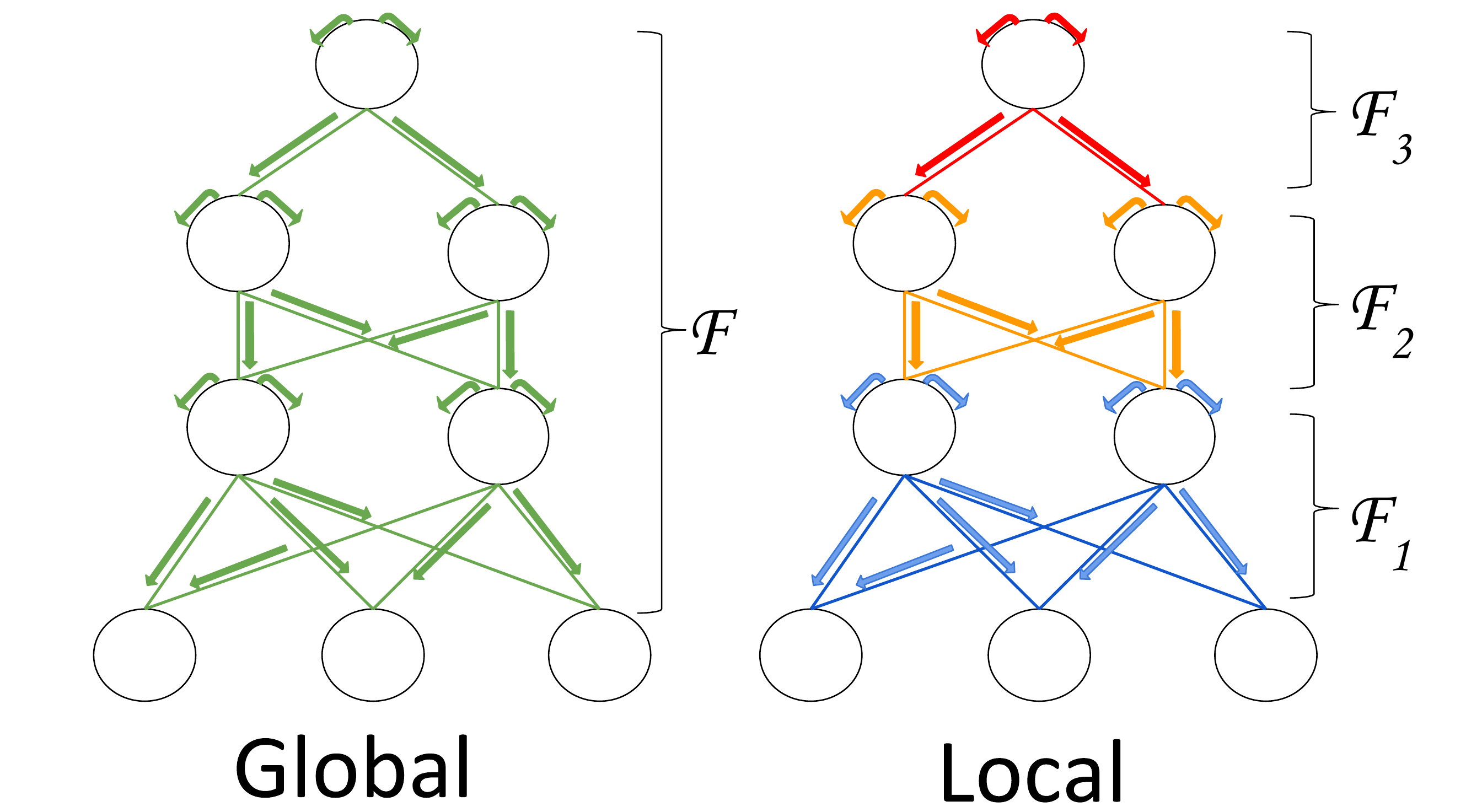}
  \caption{\footnotesize Gradients flow for computing the importance weight. Local considers the gradients of each layer independently. }
    \label{fig:local_vs_global}
  \vspace*{-30pt}
\end{wrapfigure}
In this section, we propose a local version of our method, by applying it to a single layer of the network rather than to the network as a whole. Next, we show an interesting connection between this local version and 
Hebbian learning~\cite{hebb2002organization}.

\noindent
{\bf A local version of our method.} 
Instead of considering the function $F$ that is learned by the network as a whole, we decompose it in a sequence of functions $F_l$ each corresponding to one layer of the network, i.e.,~$F(x) = F_L(F_{L-1}(...(F_1(x))))$, with $L$ the total number of layers.
By locally preserving the output of each layer given its input, we can preserve the global function $F$. This is further illustrated in Figure~\ref{fig:local_vs_global}. Note how ``local" and ``global" in this context relate to the number of layers over which the gradients are computed.

We use $y^k_i$ to denote the activation of neuron $n_i$ 
for a given input $x_k$.
Analogous to the procedure followed previously, we consider the squared $\ell_2$ norm of each layer after the activation function. %
An infinitesimal change $\delta_l = \{\delta_{ij}\}$ in the parameters $\theta_l = \{\theta_{ij}\}$ of layer $l$ %
results in a change to the squared $\ell_2$ norm  of the local function $F_l$ %
for a given input to that layer $y^k = \{y^k_i\} = F_{l-1}(...(F_1(x_k)))$ given by:
\begin{equation}
\ell^2_2( F_l(y^k;\theta_l+ \delta_l)) -\ell^2_2 (F_l(y^k;\theta_l))\approx \sum_{i,j} g_{ij}(x_k) \delta_{ij}
\end{equation}
 where $g_{ij}(x_k)=\frac{\partial[\ell^2_2( F_l(y^k;\theta_l))]}{\partial \theta_{ij}}$. In the case of a ReLU activation function, it can be shown that (see Appendix.\ref{sec:App_hebb}): 
\begin{equation}
  g_{ij}(x_k)=2*y^k_i*y^k_j
\end{equation}
Again we consider the accumulation of the gradients evaluated at different data points $\{x_k\}$ as a measure for the importance of the parameter $\theta_{ij}$:
\begin{equation}
\Omega_{ij}=\frac{1}{N}\sum_{k=1}^N g_{ij}(x_k) = 2*\frac{1}{N}\sum_{k=1}^N y^k_i*y^k_j
\label{eq:local}
\end{equation}

\noindent
{\bf Link with Hebbian theory.} 
In neuroscience, Hebbian learning theory~\cite{hebb2002organization} provides an explanation for the phenomenon of synaptic plasticity. It postulates that
``cells that fire together, wire together'': the synapses (connections) between neurons that fire synchronously for a given input
are strengthened over time to maintain and possibly improve the corresponding outputs. 
Here we reconsider this theory from the perspective of an artificial neural network after it has been trained successfully with backpropagation. 
Following Hebb's rule, parameters connecting neurons that often fire together (high activations for both, i.e. highly correlated outputs) %
are more important for the given task than those that fire asynchronously or with low activations. 
As such, the importance weight $\Omega_{ij}$ for the parameter $\theta_{ij}$ %
can be measured purely locally in terms of the correlation between the neurons' activations, i.e. 
\begin{equation}
\Omega_{ij}=\frac{1}{N}\sum_{k=1}^N y^k_i*y^k_j
\label{eq:hebb}
\end{equation}
The similarity with equation~\ref{eq:local} is striking. We can conclude that applying Hebb's rule to measure the importance of the parameters in a neural network can be seen as a local variant of our method that considers 
only one layer at a time instead of 
the global function learned by the network.
Since only the relative importance weights really matter, the scale factor $2$ can be ignored. \\ %

\noindent
\subsection{Discussion} \label{subsec:discussion}
Our global and local methods both have the advantage of computing the importance of the parameters on any given data point without the need to access the labels or the condition of being computed while training the model.
The global version needs to compute the gradients of the output function while the local variant (Hebbian based) can be computed locally by multiplying the input with the output of the connecting neurons. 
{\it Our proposed method (both the local and global version) resembles an implicit memory included for each parameter of the network. 
We, therefore, refer to it as Memory Aware Synapses. }
It keeps updating its value based on the activations of the network when applied to new data points. It can adapt and specialize to a given subset of data points rather than preserving every functionality in the network. Further, the method can be added after the network is trained. It can be applied on top of any pretrained network and compute the importance on any set of data without the need to have the labels. This is an important criterion that differentiates our work from methods that rely on the loss function to compute the importance of the parameters.

\section{Experiments}\label{sec:experiments}
We start by comparing our method to different existing LLL methods in the standard sequential learning setup of object recognition tasks. We  further analyze the behavior and some design choices of our method. 
Next, we move to the more challenging problem of continual learning of $<$subject, predicate, object$>$ triplets in an embedding space (section~\ref{sec:facts}).

\subsection{Object Recognition}
\label{sec:objrec} 

We follow the standard setup commonly used in  computer vision  to evaluate LLL methods~\cite{aljundi2016expert,li2016learning,rannen2017encoder}. It consists of a sequence of supervised classification tasks each from a particular dataset. 
Note that this  assumes having different classification layers for each task (different ``heads") that remain unshared.
Moreover, an oracle is used at test time to decide on the task (\ie, which classification layer to use). \newline
\noindent
{\bf Compared Methods.} 
\noindent
- {\em Finetuning} ({\tt FineTune}).  After learning the first task and when receiving a new task to learn,  the parameters of the network are finetuned on the new task data. This baseline is expected to suffer from forgetting the old tasks while being advantageous for the new task.

\noindent
- {\em Learning without Forgetting~\cite{li2016learning}} ({\tt LwF}). Given a new task data, the method records the probabilities obtained from the previous tasks heads and uses them as targets when learning a new task in a surrogate loss function. To further control the forgetting, the method relies on first training the new task head while freezing the shared parameters as a warmup phase and then training all the parameters until convergence.

\noindent
- {\em Encoder Based Lifelong Learning~\cite{rannen2017encoder}} ({\tt EBLL}) builds on LwF and learns a shallow encoder  on the features of each task. A penalty on the changes to the encoded features  accompanied with the distillation loss is  applied to reduce the forgetting of the previous tasks. Similar to LwF, a warmup phase is used before the actual training phase.

\noindent
- {\em Incremental Moment Matching ~\cite{lee2017overcoming}} ({\tt IMM}). A new task is learned with an L2 penalty equally applied  to the changes to the shared parameters. At the end of the sequence, the obtained models are merged through a  first or second moment matching. In our experiments, mean {\tt IMM} gives better results on the two tasks experiments while mode {\tt IMM} wins on the longer sequence. Thus, we report the best alternative in each experiment.

\noindent
- {\em Elastic Weight Consolidation~\cite{kirkpatrick2016overcoming}} ({\tt EWC}). 
It  is the first work that suggests regularizing the network parameters while learning a new task using as importance measure the diagonal of the Fisher information matrix. EWC uses individual penalty for each  previous task,  however, to make it computationally feasible  we  apply a single penalty as pointed out by~\cite{Huszar201717042}. Hence, we use a running sum of the Fishers in the 8 tasks sequence.

\noindent
- {\em Synaptic Intelligence~\cite{zenke2017improved}} ({\tt SI}). 
This method shows state-of-the-art performance and comes closest to our approach. It estimates the importance weights in an online manner while training for a new task. Similar to {\tt EWC} and our method changes to  parameters important for previous tasks are penalized during training of later tasks.

\noindent
- {\em Memory Aware Synapses } ({\tt MAS}). Unless stated otherwise, we use the global version of our method and with the importance weights estimated only on training data. We use a regularization parameter $\lambda$ of 1; note that no tuning of $\lambda$ was performed as we assume no access to previous task data.\newline 
\noindent
{\bf Experimental setup.}  
We use the AlexNet~\cite{NIPS2012_4824} architecture pretrained on Imagenet~\cite{ILSVRC15} from~\cite{krizhevsky2014one}\footnote{We use the pretrained model available in Pytorch. Note that it differs slightly from other implementations used \eg in~\cite{li2016learning}. }.
All the training of the different tasks have been done with stochastic gradient descent for 100 epochs and a batch size of 200 using the same learning rate as in~\cite{aljundi2016expert}.  Performance is measured in terms of classification accuracy. \newline
\noindent
{\bf Two tasks experiments.}  
We first consider sequences of two tasks based on three datasets:  MIT \textit{Scenes} \cite{quattoni2009recognizing} for indoor scene classification (5,360 samples), Caltech-UCSD \textit{Birds}~\cite{WelinderEtal2010} for fine-grained bird classification (5,994 samples), and Oxford \textit{Flowers}~\cite{Nilsback08} for fine-grained flower classification (2,040 samples). We consider: Scene $\rightarrow$ Birds, Birds$\rightarrow$ Scenes, Flower$\rightarrow$Scenes and Flower$\rightarrow$  Birds,  as used previously in~\cite{aljundi2016expert,li2016learning,rannen2017encoder}. We didn't consider Imagenet as a task in the sequence as this would require retraining the network from scratch to get the importance weights for {\tt SI}.
\begin{table*}[t]
\centering
\tabcolsep=0.11cm
\footnotesize
\resizebox{\textwidth}{!}{
\begin{tabular}{|l||c|c||c|c||c|c||c|c||c|c|}
\hline
Method&\multicolumn{2}{c||}{Birds $\rightarrow$ Scenes} &\multicolumn{2}{c||}{Scenes $\rightarrow$ Birds}&\multicolumn{2}{c||}{Flower $\rightarrow$ Birds}&\multicolumn{2}{c||}{Flower $\rightarrow$ Scenes}\\ \hline
{\tt FineTune} &{45.20~(-8.0)} &\bf{57.8}& {49.7~(-9.3)} &\bf{52.8}& {64.87~(-13.2)} &\bf{53.8}& {70.17~(-7.9)} &{57.31}\\ 
{\tt LwF}~\cite{li2016learning} 
&{51.65~(-2.0)} &{55.59}& {55.89~(-3.1)} &{49.46}& {73.97~(-4.1)} & {53.64}& {76.20~(-1.9)} &{58.05} \\ 
{\tt EBLL}~\cite{rannen2017encoder} 
&{52.79~(-0.8)} &{55.67}& {56.34~(-2.7)} &{49.41}& {75.45~(-2.6)} & {50.51}& {76.20~(-1.9)} &\bf{58.35} \\ 
{\tt IMM}~\cite{lee2017overcoming} 
&{51.51~(-2.1)} &{52.62}& {54.76~(-4.2)} &{52.20}& {75.68~(-2.4)} & {48.32}& {76.28~(-1.8)} &{55.64} \\ 
{\tt EWC}~\cite{kirkpatrick2016overcoming} 
&{52.19~(-1.4)} &{55.74}& \bf{58.28~(-0.8)} &{49.65}& {76.46~(-1.6)} & {50.7}& {77.0~(-1.1)} &{57.53} \\ 
{\tt SI}~\cite{zenke2017improved} %
&{52.64~(-1.0)} &{55.89}& {57.46~(-1.5)} &{49.70}& {75.19~(-2.9)} & {51.20}& {76.61~(-1.5)} &{57.53} \\ 
{\tt MAS} (ours) &\bf{53.24~(-0.4)} &{55.0}& {57.61~(-1.4)} &{49.62}& \bf{77.33~(-0.7)} &{50.39}& \bf{77.24~(-0.8)} &{57.38} \\ \hline
\end{tabular}}
\caption{\footnotesize Classification accuracy (\%), drop in first task (\%) for various sequences of 2 tasks using the object recognition setup. 
}%
\label{tab:res_twotasks_classification}
\vspace*{-20pt}
\end{table*}
As shown in Table~\ref{tab:res_twotasks_classification},
{\tt FineTune} %
clearly 
suffers from catastrophic forgetting with a drop in performance from $8\%$ to $13\%$. All the considered methods manage to reduce the forgetting over fine-tuning significantly while having performance close to fine-tuning on the new task.  
On average, our method method achieves the lowest forgetting rates (around $1\%$) while performance on the new task is almost similar ($0 - 3\%$ lower). 
\newline
{\bf Local vs. global MAS on training/test data.} 
Next we analyze the performance of our method when preserving the global function learned by the network after each task ({\tt MAS}) and its local Hebbian-inspired variant described in section~\ref{subsec:hebb} ({\tt l-MAS}). We also evaluate our methods, {\tt MAS} and {\tt l-MAS}, when using unlabeled test data and/or labeled training data.
Table~\ref{tab:res_twotasks_classification_train_test} shows, independent from the set used for computing the importance of the weights, for both {\tt l-MAS} and {\tt MAS} the preservation of the previous task and the performance on the current task are quite similar. This illustrates  our method ability to estimate the parameters importance of a given task given any set of points, without the need for labeled data. Further, computing the gradients locally at each layer for {\tt l-MAS} allows for faster computations but less accurate estimations. As such,   {\tt l-MAS} shows an average forgetting of $3\%$ compared to $1\%$ by {\tt MAS}.%
\newline
{ \bf $\ell^2_2$ vs. vector output.} We explained in section \ref{sec:method} that considering the gradients of the learned function to estimate the parameters importance would require as many backward passes as the length of the output vector. To avoid this complexity, we suggest using the square of the $\ell_2$ norm of the function to get a scalar output. We run two experiments, Flower$\rightarrow$Scenes and Flower$\rightarrow$  Birds once with computing the gradients with respect to the vector output and once with respect to the $\ell^2_2$ norm. We observe no significant difference  on forgetting over 3 random trials  where we get a mean, over 6 numbers, of $ 0.51\% \pm 0.18$  for the drop on the first task in the vector output case compared to $0.50\% \pm 0.19$ for the $\ell^2_2$ norm case.  No significant difference is observed on the second task either. As such, using $\ell^2_2$  is $n$ times faster (where $n$ is the length of the output vector)  without loss in performance.
\newline
\begin{table*}[t]
\centering
\footnotesize
\resizebox{\textwidth}{!}{
\begin{tabular}{|l|l||c|c||c|c||c|c||c|c||c|c|}
\hline
Method&$\Omega_{ij}$ computed. on&\multicolumn{2}{c||}{Birds $\rightarrow$ Scenes} &\multicolumn{2}{c||}{Scenes $\rightarrow$ Birds}&\multicolumn{2}{c||}{Flower $\rightarrow$ Bird}&\multicolumn{2}{c||}{Flower $\rightarrow$ Scenes}\\ \hline
{\tt MAS} & Train&{53.24~(-0.4)} &{55.0}& {57.61~(-1.4)} &{49.62}& {77.33~(-0.7)} &{50.39}& {77.24~(-0.8)} &{57.38} \\ 
{\tt MAS} & Test& {53.43~(-0.2)} &{55.07}& {57.31~(-1.7)} &{49.01} &{77.62~(-0.5)} &{50.29}& {77.45~(-0.6)} &{57.45} \\ 
{\tt MAS}& Train + Test& {53.29~(-0.3)} &{56.04}& {57.83~(-1.2)} &{49.56}& {77.52~(-0.6)} &{49.70}& { 77.54~(-0.5)} &{57.39} \\ \hline
{\tt l-MAS} & Train& {51.36~(-2.3)} &{55.67}& {56.79~(-2.2)} &{49.08}& {73.96~(-4.1)} &{50.5}& {76.20~(-1.9)} &{56.68} \\ 
{\tt l-MAS} & Test& {51.62~(-2.0)} &{53.95}& {55.74~(-3.3)} &{50.43}& {74.48~(-3.6)} &{50.32}& {76.56~(-1.5)} &{57.83} \\ 
{\tt l-MAS} & Train + Test& {52.15~(-1.5)} &{ 54.40}& {56.79~(-2.2)} &{48.92}& {73.73~(-4.3)} &{50.5}& { 76.41~(-1.7)} &{57.91} \\ \hline
\end{tabular}}
\caption{\footnotesize Classification accuracies (\%) for the object recognition setup - comparison between using Train and Test data (unlabeled) to compute the  parameter importance $\Omega_{ij}$.}
\label{tab:res_twotasks_classification_train_test}
\vspace*{-30pt}
\end{table*}

\noindent
{\bf Longer Sequence\ }
\newline
\begin{wrapfigure}[12]{r}{0.4\textwidth}
\vspace*{-25pt}
\centering
\small
  \includegraphics[width=0.4\textwidth]{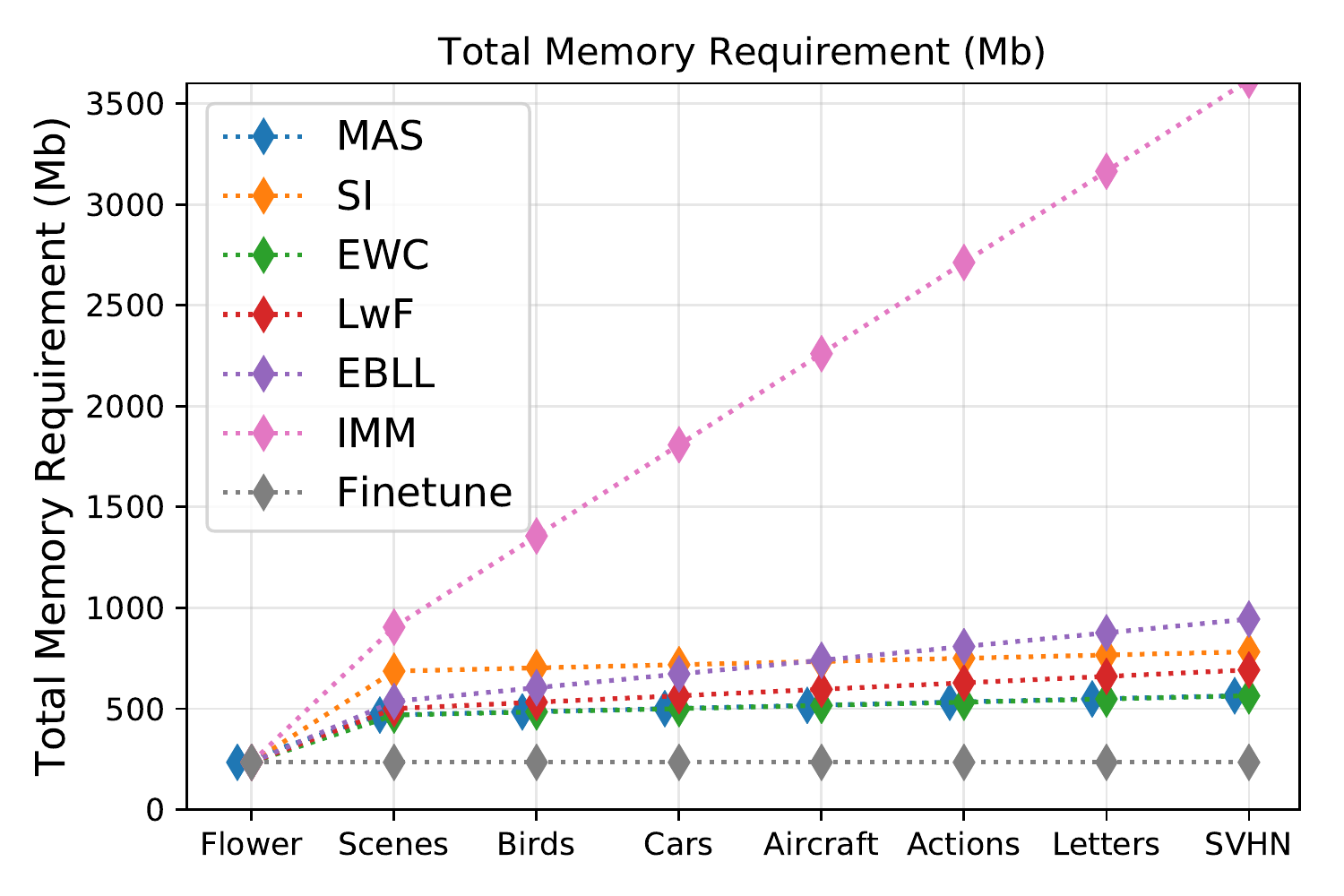}
  \caption{\footnotesize  Overall memory requirement for each method at each step of the sequence.}
   \label{fig:memreq}
\end{wrapfigure}
\begin{figure}[t]
  \centering
  \subfloat[Subfigure 1 list of figures text][\footnotesize]{
    \includegraphics[width=0.5\textwidth]{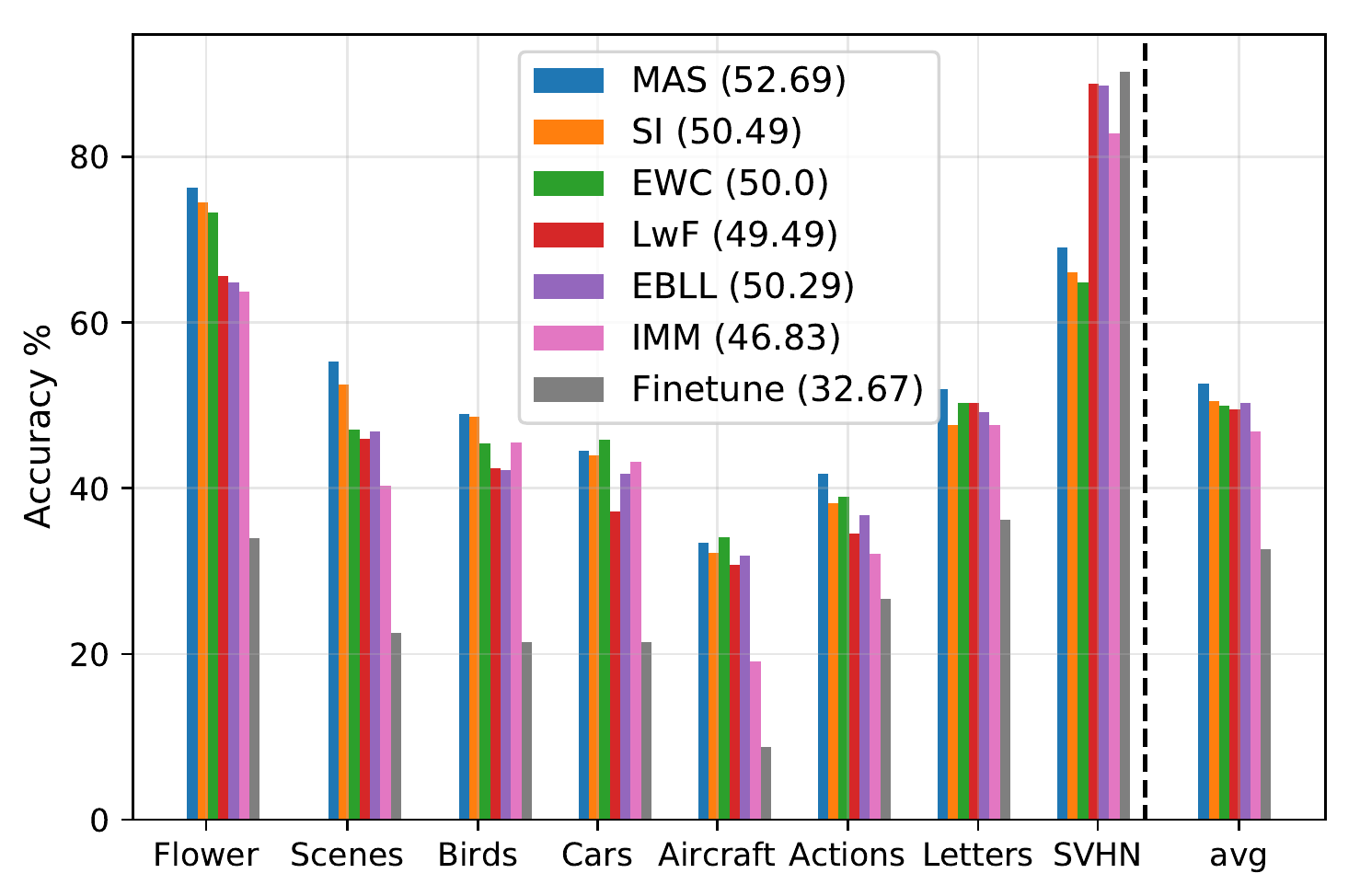}
    \label{fig:8task_acc}}
    \hfillx
  \subfloat[Subfigure 2 list of figures text][\footnotesize]{
    \includegraphics[width=0.49\textwidth]{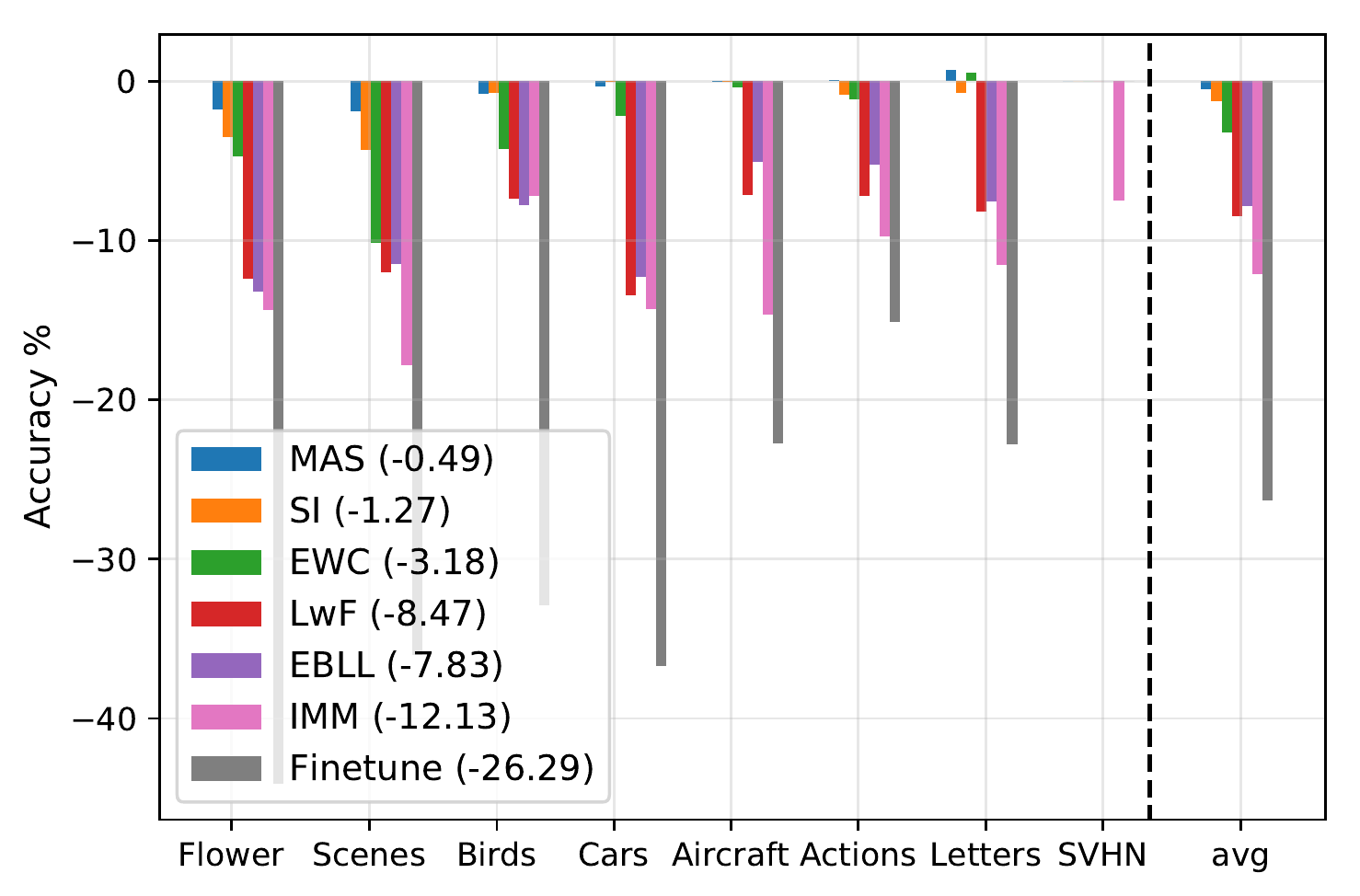}
    \label{fig:8task_drop}}
  \vspace*{-3mm}
  \caption{ \footnotesize  ~\ref{fig:8task_acc} performance  on each task, in accuracy, at the end of 8 tasks object recognition sequence.~\ref{fig:8task_drop} drop in each task relative to the performance achieved after training each task.  }
  \vspace{-20pt}
  \label{fig:8task}
\end{figure}
While the two tasks setup gives a detailed look at the average expected forgetting when learning a new task, it remains easy. Thus, we next consider a sequence of 8 tasks.
\newline
To do so, we add five more datasets: Stanford  {\tt Cars}~\cite{krause20133d} for fine-grained car classification; FGVC-{\tt Aircraft}~\cite{maji13fine-grained} for fine-grained aircraft classification; VOC {\tt Actions}, the human  action classification subset of the VOC  challenge 2012~\cite{pascal-voc-2012}; {\tt Letters}, the Chars74K dataset 
\cite{deCampos09} for character recognition in natural images; and the Google Street View House Numbers {\tt SVHN} dataset~\cite{netzer2011reading} for digit recognition. 
\newline
Those datasets were also used in~\cite{aljundi2016expert}. We run the different methods on the following sequence:  
Flower$\rightarrow$Scenes$\rightarrow$Birds$\rightarrow$Cars$\rightarrow$Aircraft$\rightarrow$Actions$\rightarrow$Letters$\rightarrow$SVHN.\newline
While Figure~\ref{fig:8task_acc} shows the performance on each task at the end of the sequence,~\ref{fig:8task_drop} shows the observed forgetting on each task at the end of the sequence (relative to the performance right after training that task). 
The differences between the compared methods become more outspoken.
{\tt Finetuning} suffers from a severe forgetting on the previous tasks while being advantageous for the last task, as expected. {\tt LwF}~\cite{li2016learning} suffers from a buildup of errors when facing a long sequence while {\tt EBLL}~\cite{rannen2017encoder} reduces slightly this effect. {\tt IMM}~\cite{lee2017overcoming} merges the models at the end of the sequence and the drop in performance differs between tasks.
More importantly,  the method performance on the last task is highly affected by the moment matching. {\tt SI}~\cite{zenke2017improved} followed by {\tt EWC}~\cite{kirkpatrick2016overcoming} has the least forgetting among our methods competitors. {\tt MAS}, our method, shows a minimal or no forgetting on the different tasks in the sequence with an average forgetting of $0.49\%$. It is worth noting that our method's absolute performance on average including the last task is $2\%$ better than {\tt SI} which indicates our method ability to accurately estimate the importance weights and the new tasks to adjust accordingly.  
Apart from evaluating forgetting, we analyze the memory requirements of each of the compared methods. Figure~\ref{fig:memreq} illustrates the memory usage of each method at each learning step in the sequence. After {\tt Finetune} that doesn't treat forgetting, our method has the least amount of memory consumption. Note that {\tt IMM} grows linearly in storage, but at inference time it only uses the obtained model.
More details on memory requirements and absolute performances, in numbers, achieved by each method can be found in the Appendix~\ref{sec:exp_details} .

\vspace{-30pt}\myparagraph{Sensitivity to the hyper parameter.}
Our method needs one extra hyper parameter, $\lambda$,
\begin{wrapfigure}[12]{r}{0.4\textwidth}
\begin{center}
\vspace{-30pt}
 \includegraphics[width=0.4\textwidth]{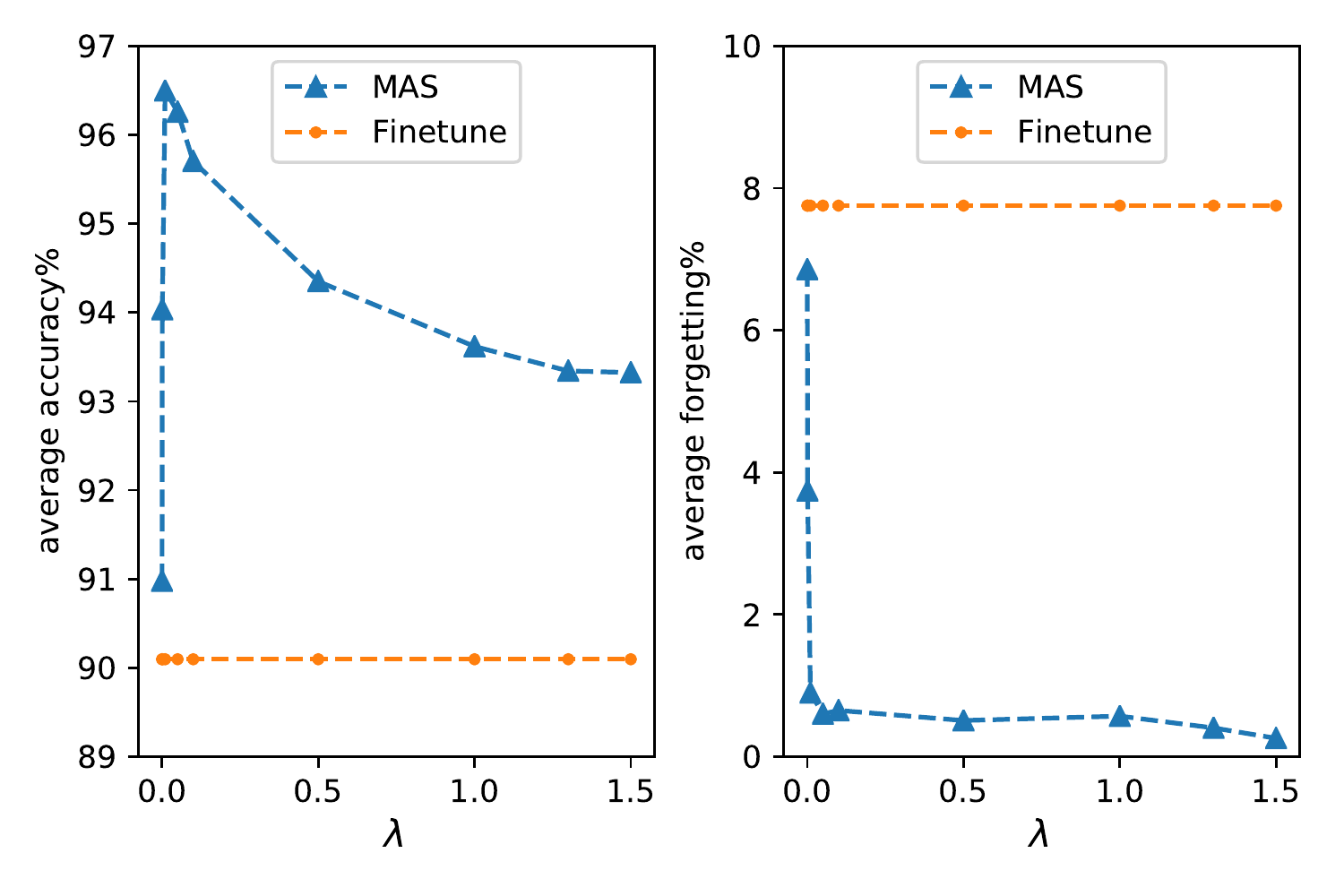}
  \caption{\footnotesize  avg. performance, left, and avg. forgetting, right, on permuted mnist sequence.}
   \label{fig:permuted}
   \end{center}
   \end{wrapfigure}
   that weights the penalty on the parameters changes as shown in Eq \ref{eq:3}.
$\lambda$ is a trade-off between the allowed forgetting and the new task loss. We set $\lambda$ to the largest value that allows an acceptable performance on the new task. For MAS, we used $\lambda=1$ in all object recognition experiments while for {\tt SI}\cite{zenke2017improved} and {\tt EWC}\cite{kirkpatrick2016overcoming} we had to vary  $\lambda$.
Figure~\ref{fig:permuted} shows the effect of $\lambda$ on the avg. performance and the avg. forgetting in   a sequence of 5 permuted MNIST tasks with a 2 layer perceptron~(512 units).  We see the sensitivity around $\lambda=1$ is very low with low forgetting, although further improvements could be achieved. \newline
{\bf Adaptation Test.} 
As we have previously explained, {\tt MAS} has the ability to adapt the importance weights to a specific subset that has been encountered at test time in an unsupervised and online manner. To test this claim, we have selected one class from the {\tt Flower} dataset, {\tt Krishna Kamal} flower. We learn the 8 tasks sequence as above while assuming {\tt Krishna Kamal} as the only encountered class. %
Hence, importance weights are computed on that subset only. At the end of the sequence, we observe a minimal forgetting on that subset of $2\%$ compared to $8\%$ forgetting on the {\tt Flower} dataset as a whole. We also observe higher accuracies on later tasks as only changes to important parameters for that class are penalized, leaving more free capacity for remaining tasks (e.g. accuracy of $84\%$ on the last task, instead of $69\%$ without adaptation). We repeat the experiment with two other classes and obtain similar results. This clearly indicates our method ability to adapt to user specific settings and to learn what (not) to forget. 

\subsection{Facts Learning}
\label{sec:facts}
Next, we move to a more challenging setup where all the layers of the network are shared, including the last layer. Instead of learning a classifier, we learn an embedding space. For this setup, we pick the problem of Fact Learning from natural images~\cite{elhoseiny2017sherlock}. For example, a fact could be ``person eating pizza''. 
We design different experimental settings to show the ability of our method to learn what (not) to forget.

\noindent
{\bf Experimental setup.} 
We use the 6DS mid scale dataset presented in~\cite{elhoseiny2017sherlock}.
It consists of  $28,624$  images, divided equally in training and test samples belonging to  $186$ unique facts. Facts are structured into 3 units: Subject (S), Object (O) and Predicate (P). We use a CNN model based on the VGG-16 architecture~\cite{simonyan2014very} pretrained on ImageNet. The last fully connected layer forks in three final layers enabling the model to have three separated and structured outputs for Subject, Predicate and Object as in~\cite{elhoseiny2017sherlock}. The loss minimizes the pairwise distance between the visual and the language embedding. For the language embedding, the Word2vec~\cite{mikolov2013efficient} representation of the fact units is used.
To study fact learning from a lifelong perspective, we divided the dataset into tasks belonging to  different groups  of facts. 
SGD optimizer is used with a mini-batch of size 35 for 300 epochs and we use a $\lambda=5$ for our method. For evaluation, we report the fact  to image retrieval  scenario. We follow the evaluation protocol  proposed in~\cite{elhoseiny2017sherlock} and report the mean average precision (MAP). For each task, we consider retrieving the images belonging to facts from this task only. We also report the mean average precision on the whole dataset which differs from the average of the performance achieved on each task. More details can be found in the supplemental materials. We focus on the comparison between the local {\tt l-MAS} and global {\tt MAS} variants of our method and {\tt SI}~\cite{zenke2017improved}, the best performing method among the different competitors as shown in Figure~\ref{fig:8task_acc}.\newline
  \begin{minipage}{\textwidth}
  
  \begin{minipage}[b]{0.6\textwidth}
  \tabcolsep=0.11cm   
  \footnotesize
\begin{tabular}{ | l| l | c| c | c | c ||c|}
\hline
&   & \multicolumn{5}{c|}{Method evaluated on}\\
\textbf{Method} & Split & \boldmath{$T_1$} &\boldmath{$T_2$}&\boldmath{$T_3$} & \boldmath{$T_4$} & all\\  \hline
\tt{Finetune} & 1 &0.19 & 0.19 &0.28&{\bf0.71}& 0.18\\
\tt{SI}\cite{zenke2017improved}
& 1 & 0.36 & 0.32 &0.38 &{ 0.68}&0.25\\ 
							
{\tt MAS (ours)} & 1 & {\bf 0.42}& {\bf 0.37} &{\bf 0.41}&0.65&{\bf 0.29}\\  \hline  
\tt{Finetune}& 2 & {0.20} & 0.27 &0.18 &{ 0.66}&0.18\\
\tt{SI}\cite{zenke2017improved}
& 2 & 0.37 & 0.39 &0.38 &{ 0.46}&0.24\\
{\tt MAS (ours)} & 2 & {\bf 0.42}& {\bf 0.42} &{\bf 0.46}&{\bf 0.65}&{\bf 0.28}\\  \hline  

\tt{Finetune}& 3 & {   0.21}& 0.25&0.24&{0.46}&0.14\\
\tt{SI }\cite{zenke2017improved}
& 3 & {\bf   0.30}& 0.31&0.36&{0.61}&0.24\\
{\tt MAS (ours)} & 3 & {\bf  0.30}& {\bf 0.36} &{\bf 0.38}&{\bf 0.66}&{\bf  0.27}\\  \hline  

\end{tabular}
\captionof{table}{\label{tab:fact_4_tasks} \footnotesize MAP for fact learning on the  4 tasks  random split, from the 6DS dataset, at the end of the sequence. } 
    \end{minipage}
    \hfill
      \begin{minipage}[b]{0.38\textwidth}
      \vspace{10pt}
    \centering
    
    \includegraphics[width=1\textwidth]{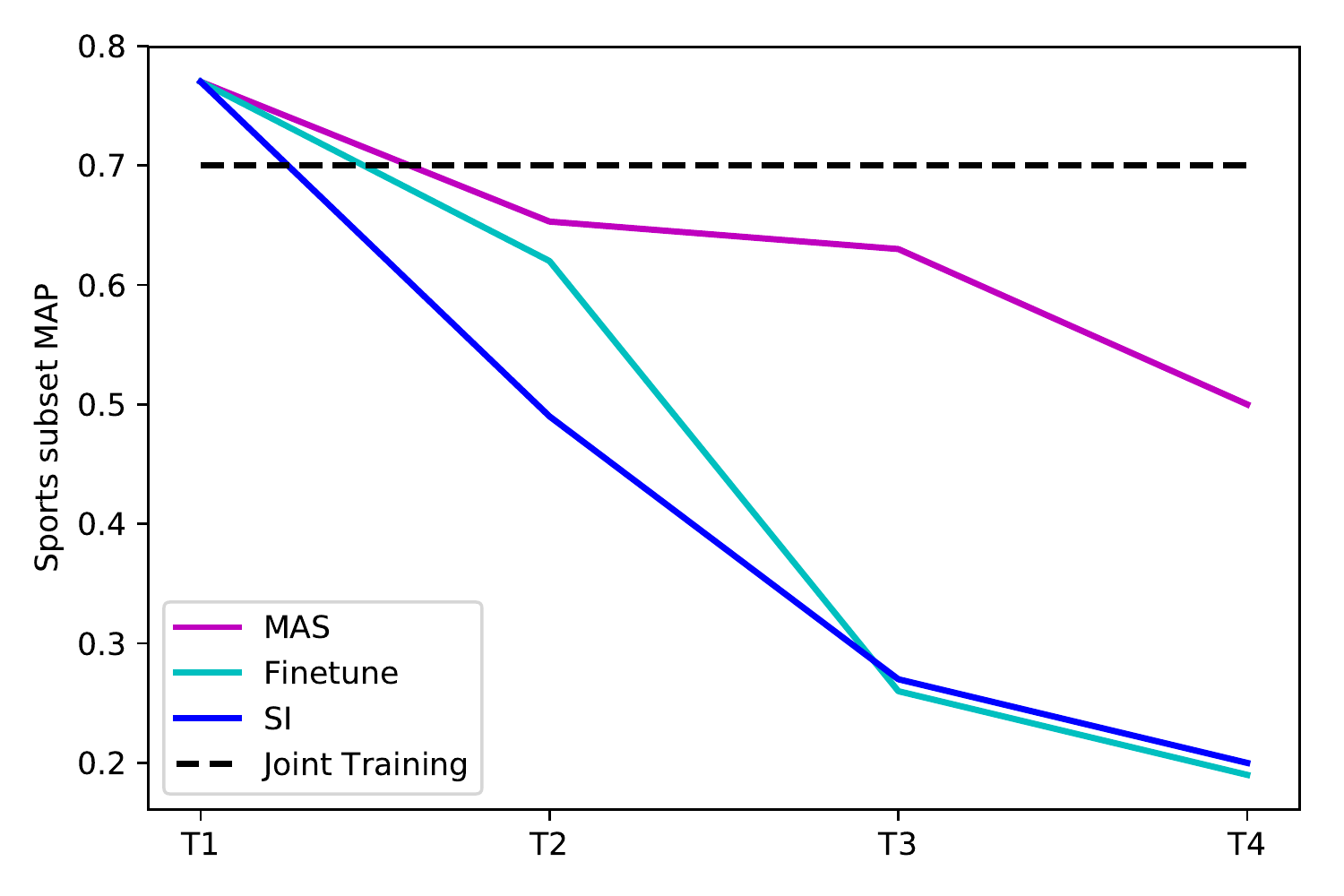}
    
    \captionof{figure}{\label{fig:adaptationx} \footnotesize   MAP on the sport subset of the 6DS dataset after each task in a 4 tasks sequence. {\tt MAS} managed to learn that the sport subset is important to preserve and prevents significantly the forgetting on this subset.}
  \end{minipage}
 \end{minipage}
 \newline
 \newline
{\bf Four tasks experiments\ }
We consider a sequence of 4 tasks obtained from randomly splitting the facts of the same dataset into 4 groups. %
Table~\ref{tab:fact_4_tasks} presents the achieved performance on each set of the 4 tasks at the end of the learned sequence based on  3 different random splits.  Similar to previous experiments, {\tt Finetune} is only advantageous on the last task while drastically suffering on the previous tasks.  However, here, our method differentiates itself clearly, showing $6$\% better MAP on the first two tasks compared to {\tt SI}. Overall, {\tt MAS} achieves a MAP of $0.29$ compared to $0.25$ by {\tt SI} and only $0.18$ by {\tt Finetune}. When {\tt MAS} importance weights are computed on both training and test data, a further improvement is achieved with  $0.30$  overall performance. This highlights our method ability to benefit from extra unlabeled data to further enhance the importance estimation.
\newline
\noindent
{\bf Adaptation Test.} 
Finally we want to test the ability of our method in learning not to forget a specific subset of a task.
When learning a new task, we care about the performance on that specific set more than the rest.
For that reason, we clustered the dataset into 4 disjoint groups of facts, representing 4 tasks, and then selected a specialized subset of \boldmath{$T_1$}, namely 7 facts of person playing sports. More details on the split can be found in the supplemental material.
We run our method with the importance parameters computed only over the examples from this set along the 4 tasks sequence.
Figure~\ref{fig:adaptationx}  shows the achieved  performance on this sport subset by each method at each step of the learning sequence. Joint Training (black dashed) is shown as reference. It violates the LLL setting as it trains on all data jointly. Note that {\tt SI} can only learn importance weights during training, and therefore cannot adapt to a particular subset. 
Our {\tt MAS} (pink) succeeds to learn that this set is important to preserve and achieves  a performance of 0.50 at the end of the sequence, 
while the performance of finetuning and {\tt SI} on this set was close to 0.20.
\section{Conclusion}  \label{sec:conculosion}
In this paper, we argued that, given a limited model capacity and unlimited evolving tasks, it is not possible to  preserve all the previous knowledge. Instead, agents should learn what (not) to forget. Forgetting should relate to the rate at which a specific piece of knowledge is used. This is similar to how biological systems are learning. In the absence of error signals, synapses connecting biological neurons strengthen or weaken based on the
concurrence of the connected neurons activations. In this work and inspired by the synaptic plasticity, we proposed a method that is able to learn the importance of network parameters from the input data that the system is active on, in an unsupervised manner. We showed that a local variant of our method can be seen  as  an application of Hebb's rule in learning the importance of parameters. We first tested our method on a sequence of object recognition problems in a traditional LLL setting. We then moved to a more challenging test case where we learn facts from images in a continuous manner. We showed i) the ability of our method to better learn the importance of the parameters using training data, test data or both;   ii)  state-of-the-art performance on all the designed experiments and iii) the ability of our method to adapt the importance of the parameters towards a frequent set of data. We believe that this is a step forward in developing systems that can always learn and adapt in a flexible manner.
\newline 
\small {\textbf{Acknowledgment:}
 The first author's PhD is funded by an FWO scholarship.

\clearpage
\appendix

\section{Additional adaptation  experiment}\label{sec:add}

\begin{figure*}
\centering
  \includegraphics[width=0.9\linewidth]{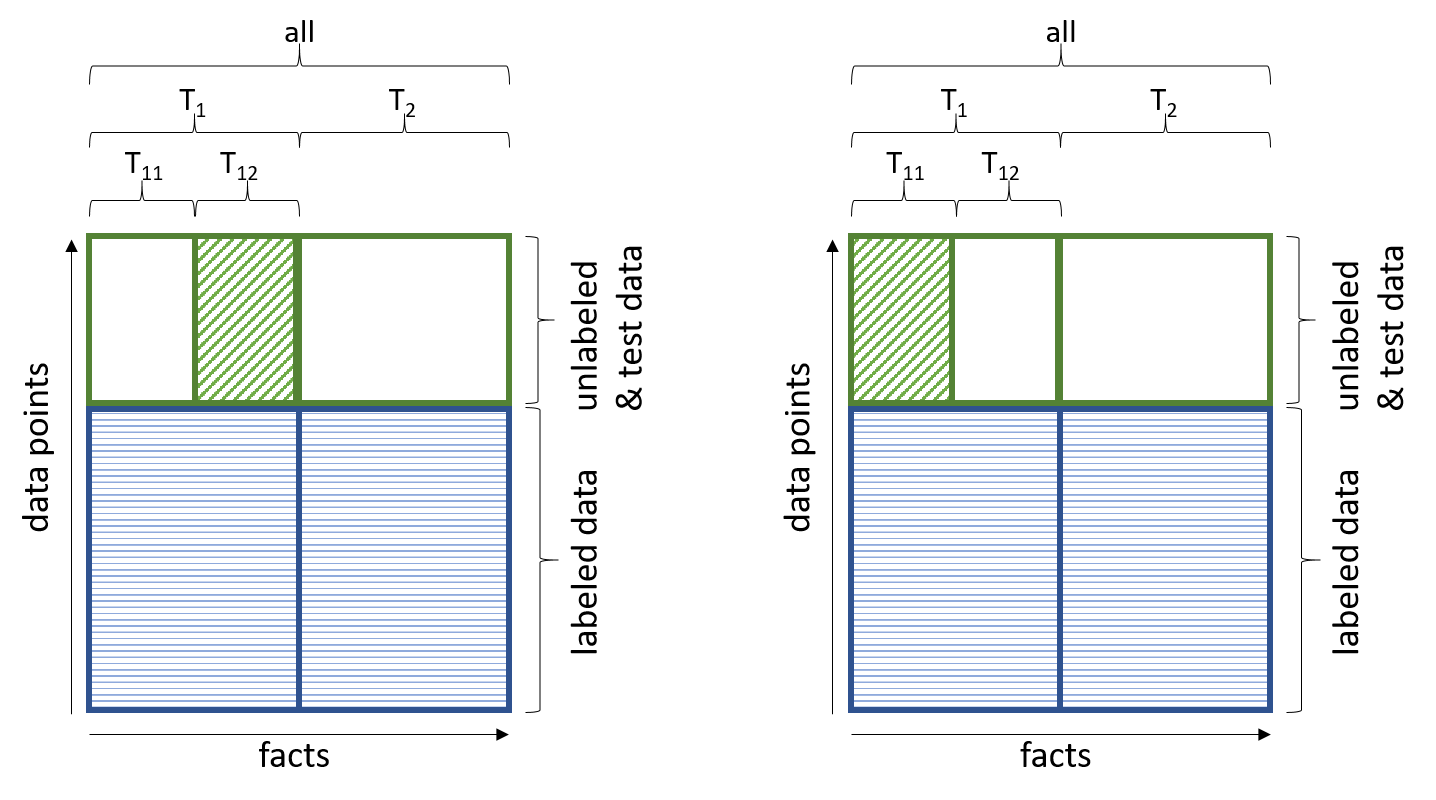}
  \caption{Visualization to clarify the training and test data splits for results of Table \ref{tab:fact_two_tasks} below. ``blue horizontal lines'' indicate training data for the model parameters $\theta_{ij}$, ``green diagonal lines'' indicate data used for computing importance weights $\Omega_{ij}$. Note that tasks are split according to facts (x-axis). Evaluation is performed for different splits on test data (green boxes, top), with the importance weights estimated based on $T_{11}$ (right) or $T_{12}$ (left). \label{fig:data-split-tab4-1}
 }

\end{figure*}

As an extra test to show that our method does not just capture general importance weights, but can really adapt, in an unsupervised fashion, to particular test conditions, we present here an additional experiment using the 6DS dataset and the two tasks experiment set up (section $5.2$ in the main paper). We split the test set of the first task data further into two random subsets of facts, $T_{11}$ and $T_{12}$. After learning the task $T_1$, the importance of the parameters is computed using one subset only ($T_{11}$ or $T_{12}$ -- we show results for both cases). Then the second task $T_2$ is learned. Figure \ref{fig:data-split-tab4-1} illustrates the designed setup for the different splits used.  Table \ref{tab:fact_two_tasks} compares the performance in each case. 

We can see that the forgetting on the subset that was used for estimating the importance of the parameters is less than on the other subset that was not considered. For example, {\tt MAS} learning the importance of the parameters on the first subset $T_{11}$ preserves a performance of $0.472$ for $T_{11}$ compared to that of $0.451$ when computing the importance on the other batch $T_{12}$. This can stand as further empirical proof of our method's capability of learning the importance of the parameters based on what the network is actively tested on.\\

\begin{table}[t]
\small
        \setlength{\tabcolsep}{3pt}
            \centering
\begin{tabular}{|l | c  |c| c || c | c |c|}
\hline
& $\Omega_{ij} $& \multicolumn{5}{c|}{Method evaluated on}\\
\textbf{Method} & comp. on & \boldmath{$T_{11}$} & \boldmath{$T_{12}$} & \boldmath{$T_1$} & \boldmath{$T_2$} & \textbf{all} \\  \hline					

{\tt l-MAS} & \boldmath{$T_{11}$ Test}& {\bf 0.389} & 0.292 & 0.251&0.508&0.272\\   
		
{\tt l-MAS} & \boldmath{$T_{12}$ Test}& 0.366& {\bf 0.319} & 0.261&0.485&0.283\\  \hline  \hline

{\tt MAS} & \boldmath{$T_{11}$ Test}& {\bf 0.472} & 0.352& 0.318 &0.475&0.307\\    
																						
{\tt MAS} & \boldmath{$T_{12}$ Test}& 0.451 & {\bf 0.361} & 0.300 &0.467& 0.297\\  \hline  
\end{tabular}
        \caption{Adaptation to test condition.  Learning  importance  weights $\Omega_{ij}$ on \boldmath{$T_{11}$} vs.~\boldmath{$T_{12}$},  two subsets of test split \boldmath{$T_1$}.  Mean average precision for fact learning, two tasks scenario, random split of 6DS dataset. Results for \boldmath{$T_{1}$}, \boldmath{$T_{2}$}, \textbf{all} as reference. Note that, for evaluation, we always use the facts of this (sub)task only, hence one should compare within column only.
         }
         \label{tab:fact_two_tasks}
         \vspace*{-0.3cm}
    \end{table}

\section{Connection to Hebbian Learning}\label{sec:App_hebb}
In this section we provide more details for our derivation in Section 4.3 in the main paper, which shows the connection of our method to Hebbian Learning.
As we explained in the main paper, 
for the local version of our method, {\tt l-MAS}, instead of considering the function $F$ that is learned by the network as a whole, we decompose it in a sequence of functions $F_l$, each corresponding to one layer of the network, i.e.~$F(x_k) = F_L(F_{L-1}(...(F_1(x_k))))$, with $L$ the total number of layers. To simplify the notations, we will drop the index $k$ referring to the input sample $x_k$ from now on.
Rather than preserving the global function $F$, {\tt l-MAS} preserves the output of each layer given its input. Here, we use $y^l_i$ to denote the activation of neuron $n_i$ at layer $l$ 
for a given input $x$. Note that, for clarity, we explicitly add a superscript $l$ referring to the layer.
Hence $y^l= \{y^l_j\}$ is the output of layer $l$ and $y^{l-1} = \{y^{l-1}_i\}$ is its input. $y^l$ is a vector with elements $\{y^l_j\}$; we use a similar notation in the paper and in the following. We use indices $i$ and $j$ for neurons of the input and output layer respectively. $\theta_{ij}$ is then a network parameter representing the connection between neuron $n_i$ in layer $l-1$ and neuron $n_j$ in layer $l$.

As explained in the main text, we consider the squared $\ell_2$ norm of each layer after the activation function. 
An infinitesimal change $\delta_l = \{\delta_{ij}\}$ 
in the parameters $\theta_l = \{\theta_{ij}\}$ of layer $l$ 
results in a change to the squared $\ell_2$ norm  of the local function $F_l$ 
for a given input to that layer 
$y^{l-1} 
= F_{l-1}(...(F_1(x)))$ given by:
\begin{equation}
\ell^2_2( F_l(y^{l-1};\theta_l+ \delta_l)) -\ell^2_2 (F_l(y^{l-1};\theta_l))\approx \sum_{i,j} g_{ij}(x) \delta_{ij}
\end{equation}
 where $g_{ij}(x)=\frac{\partial[\ell^2_2( F_l(y^{l-1};\theta_l))]}{\partial \theta_{ij}}$.
In the case of a ReLU activation function 
and considering a fully connected layer with $I*J$ parameters, i.e.
~$y^{l}_j=ReLU(out^{l}_j)$ with $out^{l}_j= \sum_{h=1}^{I}\theta_{hj}*y_h^{l-1}$, we can write
\begin{equation}
g_{ij}(x)=\frac{\partial[\ell^2_2( F_l(y^{l-1};\theta_l))]}{\partial \theta_{ij}}=\frac{\partial[\sum_{o=1}^{J} (y^{l}_o)^2]}{\partial \theta_{ij}}=\sum_{o=1}^{J} \frac{\partial(  (y^{l}_o)^2)}{\partial \theta_{ij}}
\end{equation}
Since 
$\frac{\partial ( (y^{l}_o)^2)}{\partial \theta_{ij}}=0$
when  $o \neq j$, we get
\begin{equation}
g_{ij}(x)= \frac{ \partial(  (y^{l}_j)^2)}{\partial \theta_{ij}}
= 2 *y^l_j *\frac{\partial (y^{l}_j) }{\partial \theta_{ij}}
= 2 *y^l_j *\frac{\partial (ReLU(out^{l}_j))}{\partial \theta_{ij}}
\end{equation}
§ReLU is non smooth since it has a non-linearity at 0. We show the subgradients in the two cases (also using the fact that
$out^{l}_j= \sum_{h=1}^{I}\theta_{hj}*y_h^{l-1}$): 
\begin{enumerate}

\item  if $out^{l}_j>0$, $ReLU(out^{l}_j)=out^{l}_j$ and $(ReLU)^\prime$ =1 :
\begin{eqnarray}
\frac{\partial (ReLU(out^{l}_j))}{\partial \theta_{ij}} & = &\frac{\partial(ReLU(out^{l}_j))}{\partial out^{l}_j}
\frac{\partial(out^{l}_j)}{\partial \theta_{ij}}
=\frac{\partial(out^{l}_j)}{\partial \theta_{ij}} \\
& = & \frac{\partial( \sum_{h=1}^{I}\theta_{hj}*y_h^{l-1})}{\partial \theta_{ij}} =\frac{\partial( \theta_{ij}*y_i^{l-1})}{\partial \theta_{ij}}=y_i^{l-1} \\
\Rightarrow g_{ij}(x) &= &2*y_j^l*y_i^{l-1}\label{eq:g1}
\end{eqnarray}

\item in the other case, $ReLU(out^{l}_j)=0$ and $(ReLU)^\prime=0$, so
\begin{equation}
 g_{ij}(x)=0
\end{equation}
At the same time,
\begin{equation}
 ReLU(out^{l}_j)=0 \Rightarrow y_j^l=0 \,\textrm{and} \, \, 2*y_j^l*y_i^{l-1}=0
\end{equation}
Hence
\begin{equation}\label{eq:g0}
g_{ij}(x)=2*y_j^l*y_i^{l-1}=0 
\end{equation}
\end{enumerate}
Based on equations \ref{eq:g1} and \ref{eq:g0}, we have shown
\begin{equation}
  g_{ij}(x)=2*y^{l-1}_i*y^l_j
\end{equation}
which is remarkably similar to the Hebbian learning rule as explained in the main paper.
\section{Experimental Details }\label{sec:split}

\subsection{Performance of each of the compared methods on the 8 tasks experiments}
In the main paper, we described a learning sequence composed of 8 tasks and reported the results in a bar plot (Figure~\textcolor{red}{5a} in the main paper). To ease the comparison with other methods in future work, we report the performance in accuracies achieved by each method in Table \ref{tab:8tasks}.
 \begin{table*}
 \centering
  \caption{\label{tab:8tasks} Performance  measured in accuracies achieved by each method on each of the learned tasks at the end of the 8 tasks sequence (Table for bar plot in Figure~\textcolor{red}{5a} of the main paper).}
 \begin{tabular}{ | l | c| c | c | c | c | c | c | c || c  |}
 \hline
 Method & Flower & Scenes &Birds & Cars & Aircraft & Actions & Letters & SVHN &avg \\  \hline

 {\tt Finetune}& 33.98&	22.57&	21.48&	21.48&	13.32&	26.64&	36.18&	90.33&	32.67\\  \hline  	

{\tt LwF} \cite{li2016learning} & 65.68&	46.04&	42.47&	37.24&	30.75&	34.55&	50.31&	88.89&	49.491\\  \hline 
{\tt IMM} \cite{lee2017overcoming}& 67.44 & 	47.08 & 	42.28 & 	39.19 & 	18.93 & 	32.88 & 	46.35 & 	53.15 & 	43.41\\  \hline 
{\tt SI} \cite{zenke2017improved}&  74.58&	52.53&	48.65&	43.96&	32.19&	38.22&	47.7&	66.09&	50.49\\  \hline  
									
 {\tt MAS}& 76.34	&55.3&	49.03&	44.59&	33.48&	41.76	&51.93&	69.13&	52.695	\\  \hline  

 \end{tabular}
\vspace*{-0.2cm} 
 \end{table*}

\subsection{Details on methods' memory requirements estimation}\label{sec:exp_details}
In the main paper, section~\textcolor{red}{$5.1$} Figure~\textcolor{red}{6}, we report an approximation to the memory requirements  of each  method in the 8 tasks object recognition sequence. Here, we show a detailed estimation in which we split the overall memory consumption into memory requirements at training phase (Figure \ref{fig:memory}) and storage requirements in between tasks (Figure~\ref{fig:storage}).

\begin{figure*}[h!]
\minipage{0.45\textwidth}
  \includegraphics[width=0.9\linewidth]{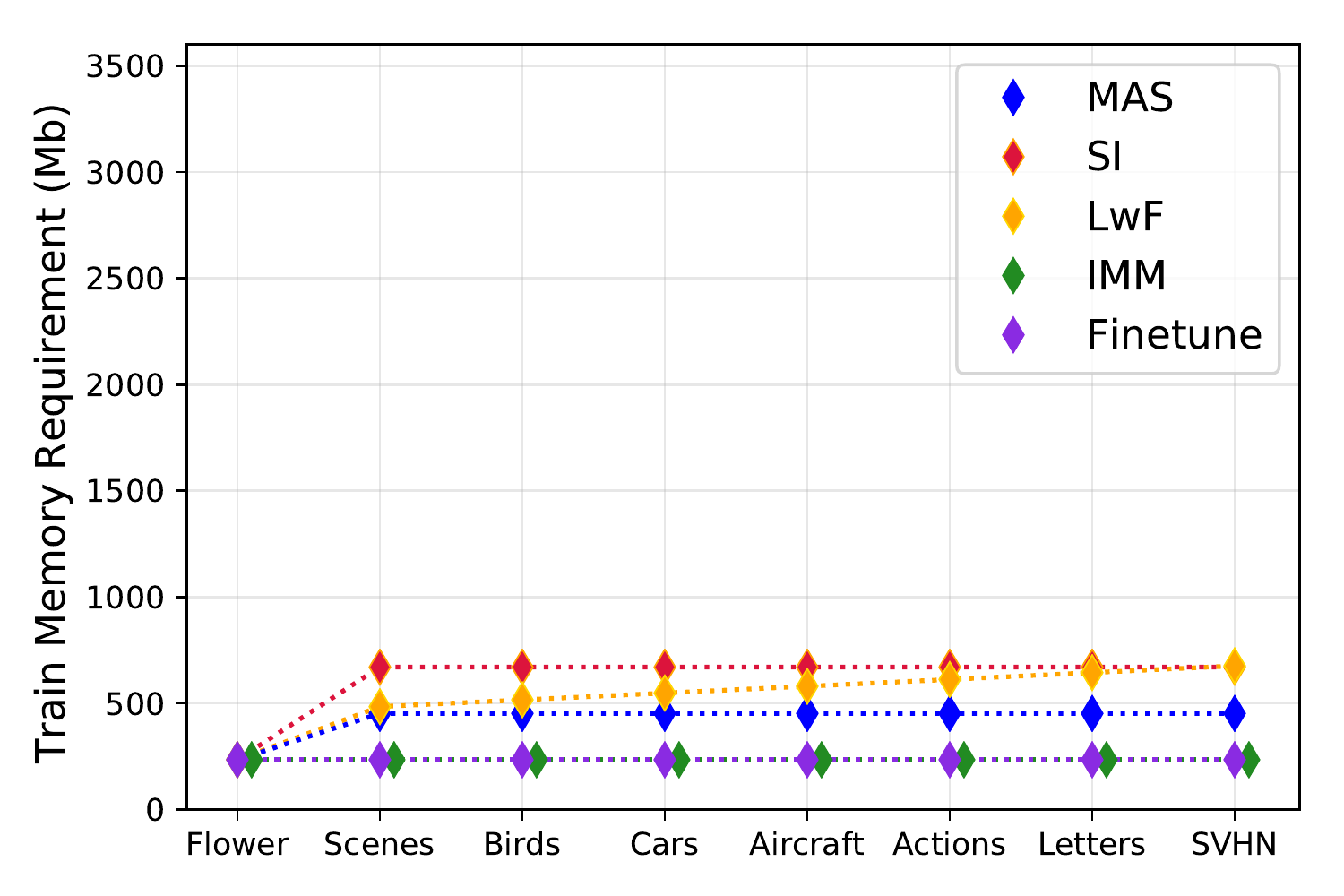}
  \caption{Memory requirements during training for each method in the 8 tasks sequence.} \label{fig:memory}
\endminipage
\hfill
  \minipage{0.45\textwidth}
  \includegraphics[width=0.9\linewidth]{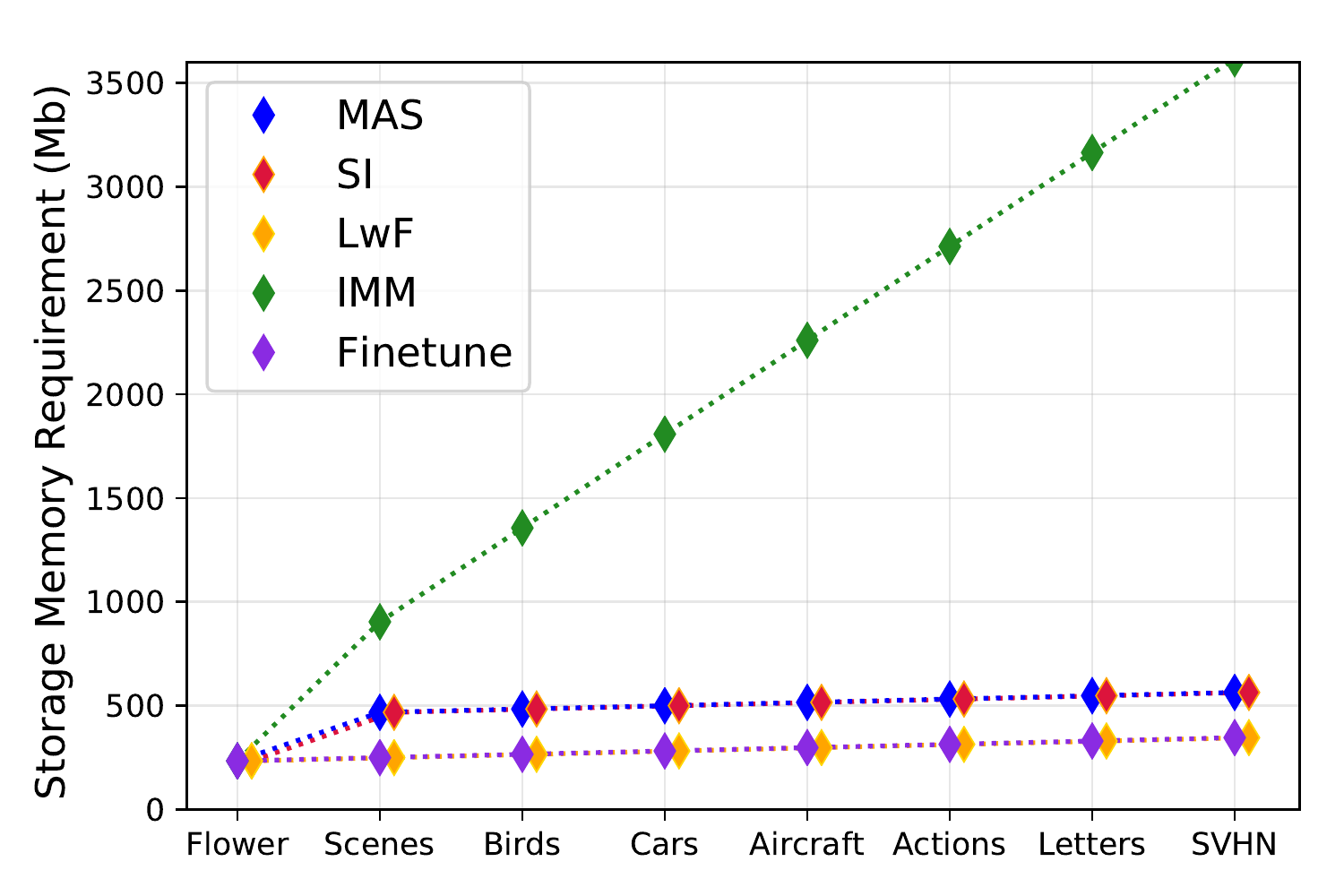}
  \caption{Storage requirements for each of the compared methods in the 8 tasks sequence.}\label{fig:storage}
\endminipage\hfill
\end{figure*}

Finetuning needs no additional memory requirements  other than the model and the tasks' heads. In spite of the fact that  {\tt IMM}~\cite{lee2017overcoming} consumes the same amount of memory as Finetuning while training, it requires an offline storage that increases linearly with the number of tasks. That's due to the storage of a Fisher information matrix for each task and previous tasks models. {\tt LwF}~\cite{li2016learning},  while training a new task, needs to load the last task model and all the previous heads to  get the target predictions. However, {\tt LwF} has the same offline storage requirements between tasks as Finetuning. {\tt SI}~\cite{zenke2017improved} and {\tt MAS} (ours) need the same offline storage requirements (importance weights  and model parameters). However, at training phase, {\tt SI} needs additional memory to accumulate the contributions of each parameter to the change in the loss.

\subsection{Fact learning experimental setup}
In the main paper, section~\textcolor{red}{5.2} paragraph "Experimental Setup", we explained the protocol followed for the Fact learning experiments. Here, we explain in more details how we get the random split for both two tasks and 4 tasks experiments. We randomly split the facts into  different groups and make sure that these groups contain a balanced number of facts and a balanced number of corresponding training and test images by selecting the best candidate out of 100 random trials. Since the "person" fact has the biggest number of images and it is a unit of a good portion of the rest of the facts, we make sure that this fact is in  the first group of facts.

\subsection{Fact learning adaptation test experimental setup}

In section~\textcolor{red}{5.2} paragraph "Adaptation Test" of the main paper, we move from the random split setup, as presented in the subsection "Longer sequence of tasks" and above, to a semantic grouping of the facts, in order to make our setup more realistic. To this end, we group the facts by their similarity in the word2vec language space.
We  use agglomerative clustering to build a dendogram  based on the Euclidean distance between the word2vec embeddings of the facts.

We selected a cut of 4 clusters.
Those clusters contain semantically related facts and can be interpreted as

\begin{itemize}
\item{\em Cluster 1:} facts describing human actions such as \fact{person, riding, bike}, \fact{person, jumping} 
\item {\em Cluster 2:} facts of different objects such as \fact{battingball}, \fact{battingstumps}, \fact{dog}, \fact{car}, etc.
\item {\em Cluster 3:} facts describing humans holding or  playing musical instruments, such as \fact{person, playing, flute}, \fact{person, 	holding, cello}, etc.
\item {\em Cluster 4:} facts describing human interactions such as \fact{person, arguing with, person}, \fact{person, dancing with, person}, etc. 
\end{itemize}

\noindent
We chose a subset of 7 facts focusing on sports:
\begin{enumerate}
\item \fact{person, batting, cricket}
\item \fact{person, bowling, cricket}
 \item \fact{person, forehead, tennis}
\item  \fact{person, play, croquet}
\item  \fact{person, serve, tennis}
\item  \fact{person, smash, volleyball}
 \item \fact{person, throwing, frisbee}
\end{enumerate}

We run our method while supposing that between the learning steps the agent is active on images belonging to these subsets and thus importance weights are computed with respect to these facts and aimed at preserving their performance.

\section{Histogram of  parameters importance $\Omega$}\label{sec:hist}
We have shown empirically  in the main paper that our proposed method ({\tt MAS}) is able to identify the important parameters and penalize  changing them when learning a new task. To  further analyze how the importance values are spread among  the different parameters, we plotted the histogram of $\Omega$ (the parameter importance). Ideally, a good importance measure would give very low importance values to the unused parameters and high values for those that are crucial for the task at hand.  

Part (a) of  Figure \ref{fig:hist_Omega} shows the histogram of $\Omega$ of the last shared convolutional layer computed on the training data from the first task.
This is based on the two tasks experiments under the fact learning setting (section 5.2). We can notice how the histogram has a peak at a value close to zero and then goes flat. Part (b) of Figure \ref{fig:hist_Omega}   shows the same histogram but magnified in the area covering the 1000 top most important parameters. We can see the long tail distribution and  how the values get sparser the more we move to higher importance assignment. This indicates that our method ({\tt MAS}) will allow changes on most of the parameters that were unused by the first task while penalizing changes to those few crucial parameters that carry meaningful information for the learned task.
\begin{figure*}[h!]
\centering

\includegraphics[width=\linewidth]{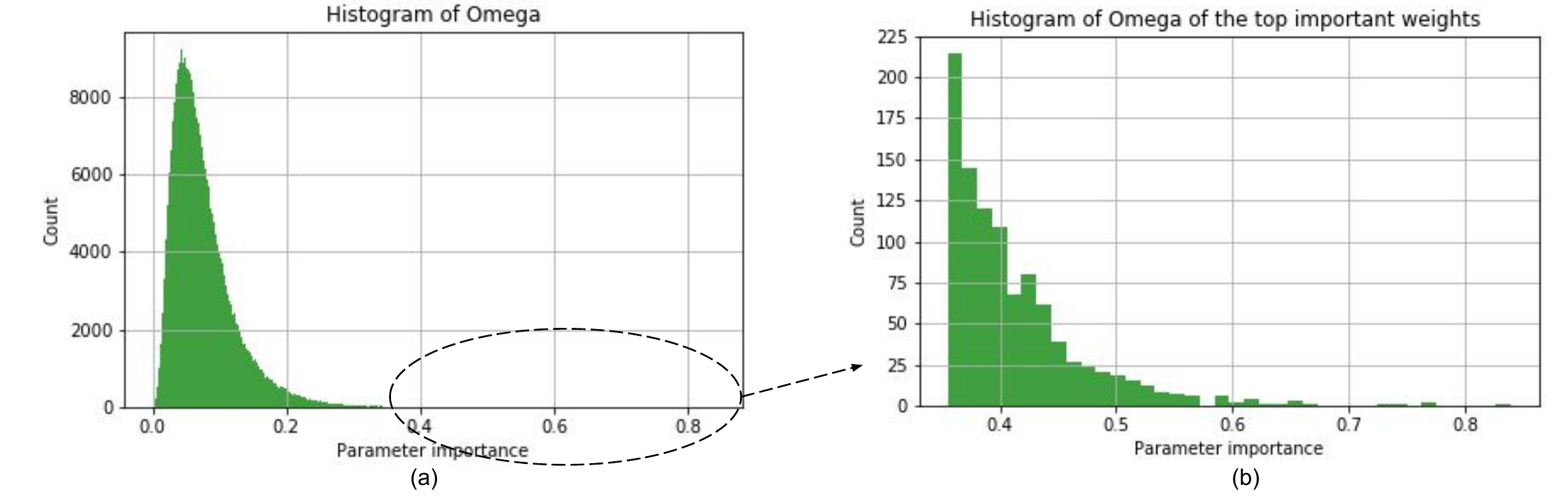}
  \caption{(a) Histogram of the parameters importance $\Omega_{ij}$ from the last shared convolutional layer, based on fact learning, two tasks experiments with $\Omega$ computed on training data. (b) A magnified look at the histogram for the top 1000 most important values. Only few important parameters have a very high value, being crucial for the specific task, while for most parameters there is a low penalty to adapt them to other tasks.}
  \label{fig:hist_Omega}

\end{figure*}

\section{Correlation between the parameters importance computed  on different sets}\label{sec:corr}
In the main paper, we have conducted experiments to examine our method's ability to preserve the previous task's performance by computing the importance of the parameters on different sets, e.g. train, test or both (see Tables~\textcolor{red}{2} and ~\textcolor{red}{4} in the main paper).  In section~\ref{sec:add}, we also have used subsets thereof.
We have shown that our method is able to adequately compute the importance of the parameters  using either the training data or the test data in an unsupervised manner. We also have shown that the method is able to adapt to a subset and preserve mostly the performance on that subset more than the rest of the task.  Here we want to investigate the correlation or the difference between the importance assigned to the parameters computed on different sets. 

\begin{figure*}[h!]

\minipage{0.45\textwidth}
  \includegraphics[width=\linewidth]{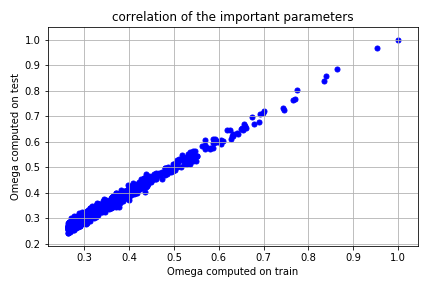}
  \caption{Top most important parameters from $\Omega$  computed on training data.  The X-axis represents the values from $\Omega$ computed on training data while the Y-axis represents the values from  $\Omega$ computed on test data.  Object recognition experiment Birds$\rightarrow$Scenes}\label{fig:cl6_train}
\endminipage\hfill
\minipage{0.45\textwidth}
  \includegraphics[width=\linewidth]{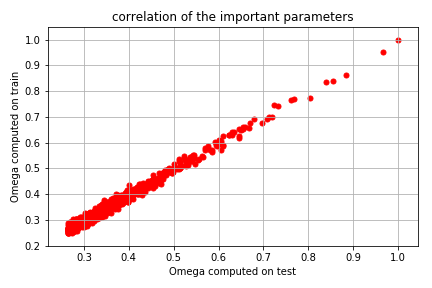}
  \caption{Top important parameters from $\Omega$ computed on test data. The X-axis represents the values from $\Omega$ computed on test data while the Y-axis represents the values from  $\Omega$ computed on training data. Based on object recognition experiment Birds$\rightarrow$Scenes}\label{fig:cl6_test}
\endminipage\hfill

\end{figure*}

First, we compare the estimated parameters importance ($\Omega$)  using the training data and the $\Omega$ computed using the test data. For that, we used a model from the object recognition experiment, namely Birds$\rightarrow$Scenes, the results of which are shown in Table~\textcolor{red}{1} in the main paper. Figure \ref{fig:cl6_train} shows a scatter plot for the top $1000$ most important parameters according to   the $\Omega$ computed on the training data (blue). The X-axis represents the values from $\Omega$ computed on training data while the Y-axis represents the values from  $\Omega$ computed on test data.
Figure \ref{fig:cl6_test} shows a similar scatter plot for the top $1000$ important parameters according to   the $\Omega$ computed on the test data~(red). Here, the X-axis represents the values from $\Omega$ computed on test data while the Y-axis represents the values from  $\Omega$ computed on training data.
A plot where the points are closely lying around a straight line indicates that the parameters from the two $\Omega$s have similar importance values. A plot where the points are spread further from  such a line and scattered among the plotted area  indicates a lower correlation between the $\Omega$s. 
It can be seen how similar are the importance values computed on test data to those computed on training data where they form a tight grouping of points around a straight line where the values would be identical. This demonstrates our method's ability to correctly identify the important parameters in an unsupervised manner, regardless of what set is used for that purpose as long as it covers the different classes or concepts of the task at hand.  \\
\begin{figure*}[h!]

\minipage{0.45\textwidth}
  \includegraphics[width=\linewidth]{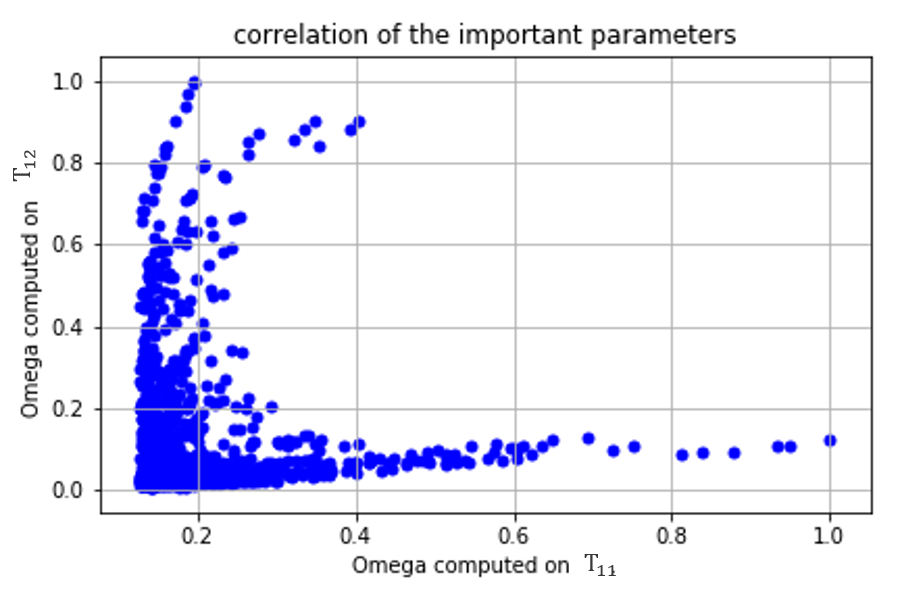}
  \caption{Top most important parameters from $\Omega$ computed on $T_{11}$. The X-axis represents the values from $\Omega$ computed on $T_{11}$ while the Y-axis represents the values from  $\Omega$ computed on $T_{12}$, under the fact learning setting, two tasks experiments. Importance shown for the last convolutional layer.}\label{fig:b11_b12_s_1}
\endminipage\hfill
\minipage{0.45\textwidth}
  \includegraphics[width=\linewidth]{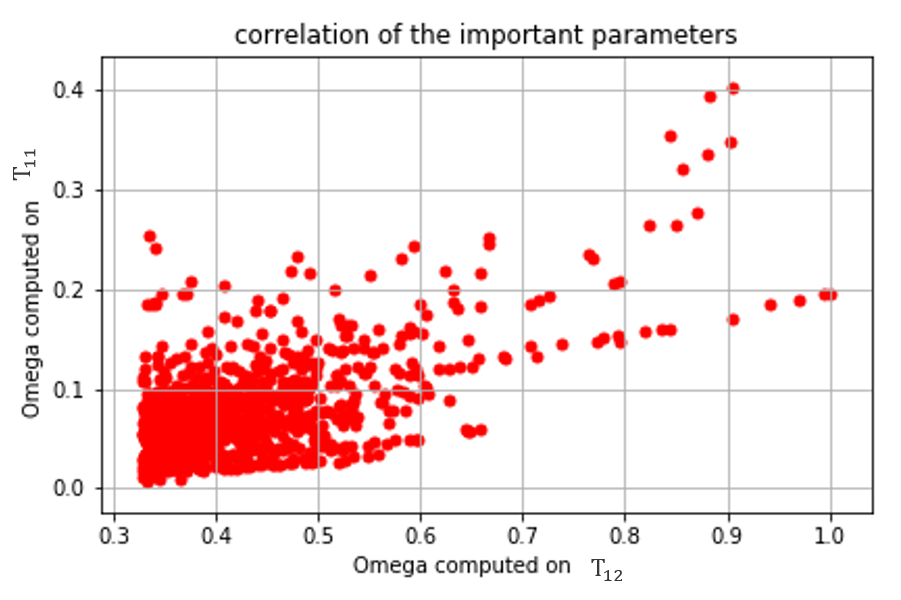}
  \caption{Top most important parameters from $\Omega$ computed on $T_{12}$. The X-axis represents the values from $\Omega$ computed on $T_{12}$ while the Y-axis represents the values from  $\Omega$ computed on $T_{11}$, under the fact learning setting, two tasks experiments \ref{tab:fact_two_tasks}. Importance shown for the last convolutional layer.}\label{fig:b11_b12_s_2}
\endminipage\hfill
\end{figure*}

How about using different subsets that cover a partial set of classes or concepts from a task? In \ref{sec:add} we have conducted an experiment under the fact learning setting where we split the data from the first task $T_1$ into two disjoint groups of facts and showed that computing the importance on one subset results in a better preservation in performance compared with the other subset that was not used for computing the importance -- see Table \ref{tab:fact_two_tasks} and Figure \ref{fig:data-split-tab4-1}. This suggests that the importance of the parameters differs while using different subsets. To further investigate  this claim, we plotted the values of $\Omega$ for the $1000$ top most important parameters estimated on the $T_{11}$ (in blue) subset of the training data  from the first task $T_1$ along with the same parameters but with their importance computed using the other subset $T_{12}$. Figures  \ref{fig:b11_b12_s_1} and \ref{fig:b11_b12_s_2} show this for the last convolutional layer.

This suggests that the method identifies the important parameters needed for each subset and when those parameters are shared the parameters importance is correlated between the two subsets while when those are different, different parameters receive different importance values based on the used subset. 

\section{Visualizing the learned embedding on the adaptation experiment }\label{sec:sport}
\begin{figure*}[ht!] 
\minipage{0.32\textwidth}
  \includegraphics[width=\linewidth]{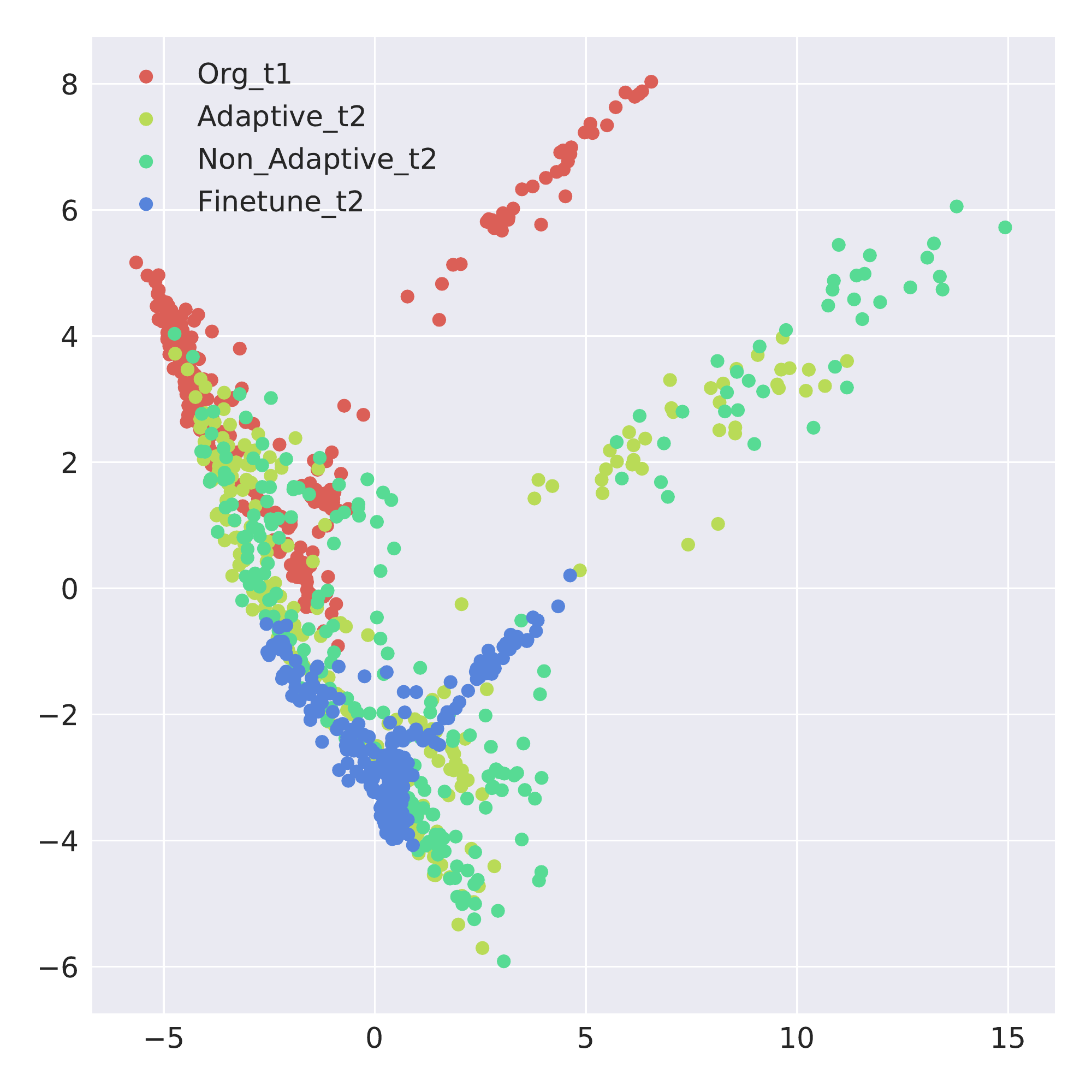}
  \caption{Projections onto a 2D embedding, after training the second task}\label{fig:t2}
\endminipage\hfill
\minipage{0.32\textwidth}
  \includegraphics[width=\linewidth]{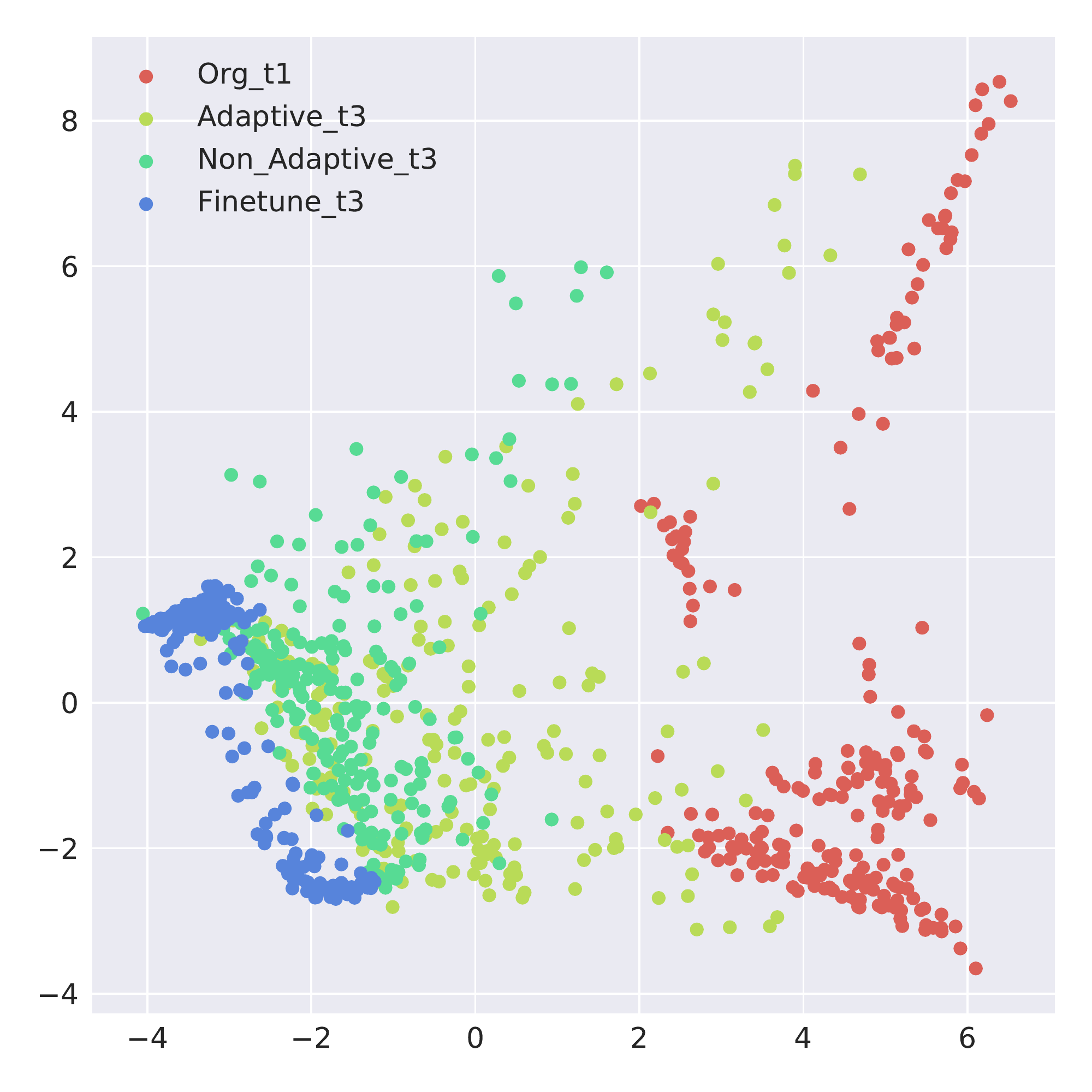}
  \caption{Projections onto a 2D embedding, after training the third task}\label{fig:t3}
\endminipage\hfill
\minipage{0.32\textwidth}%
  \includegraphics[width=\linewidth]{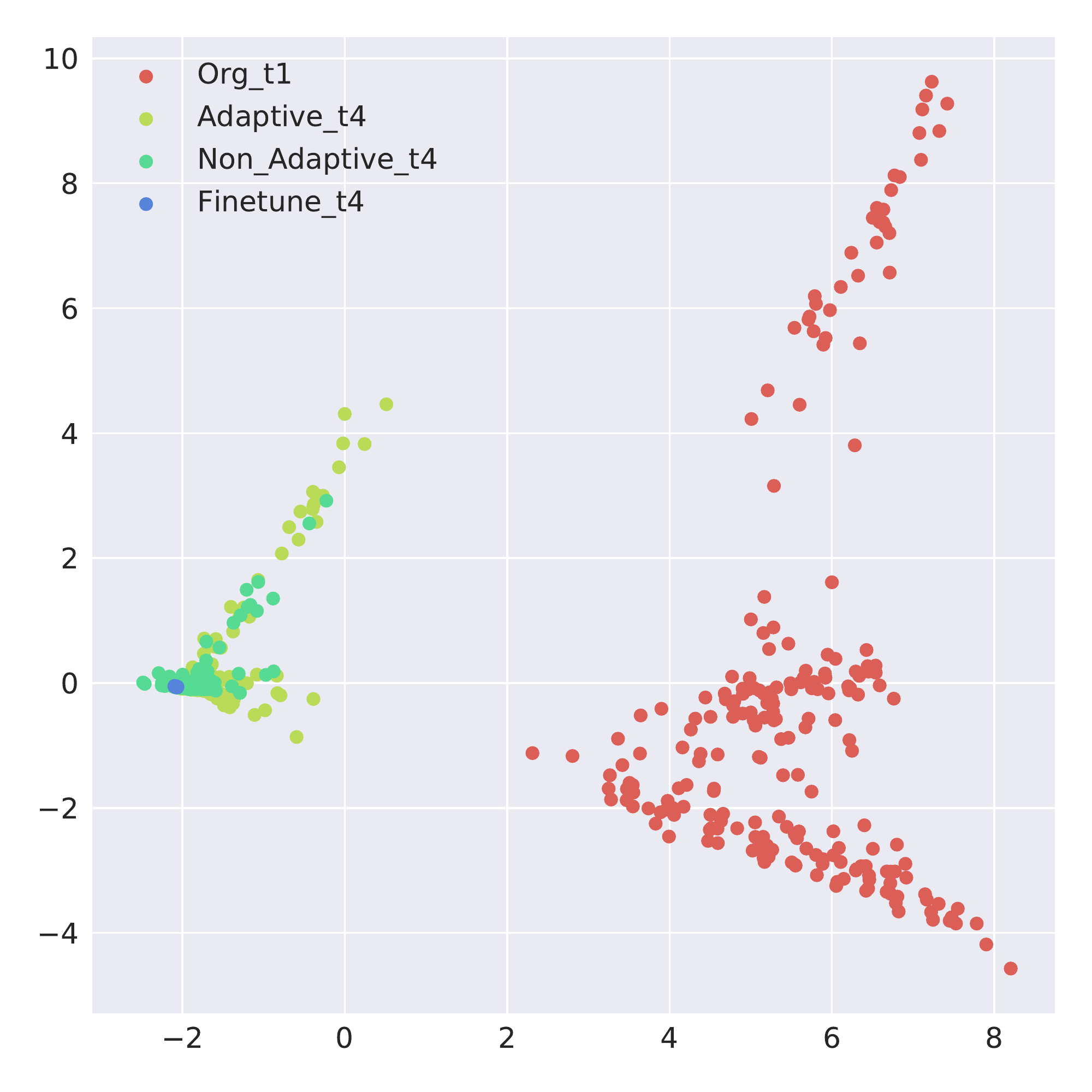}
  \caption{Projections onto a 2D embedding, after training the fourth task}\label{fig:t4}
\endminipage
\end{figure*}
Finally, in the main paper (section~\textcolor{red}{5.2}, Adaptation test paragraph), we showed that our method tries to preserve the performance on a specific subset in case  it encounters this subset frequently at test time along a learning sequence.  This has been done by picking a subset from the first task in the 4 tasks fact learning sequence. This subset was mainly composed of sports facts. We showed that our method reduces the forgetting on this subset the most among the competitors that do not have this specialization capabilities (Figure~\textcolor{red}{7} in the main paper).  We were eager to  know what happens in the learned embedding space, i.e. how the projections of the samples that belong to this subset change along the sequence compared to how they were right after training the first task. For that purpose, we extract a $2D$ projection of the learned embedding after each task in the sequence. This was done for our method ({\tt MAS}) when adapting to sport subset (\textbf{Adaptive}) and our method ({\tt MAS}) when preserving the performance on all facts of the first task (\textbf{Non Adaptive}). We also show the projections of the points in the embedding learned by the finetuning baseline (\textbf{finetune}, where no regularizer is used). To have a point of reference, we also show the projections of the originally learned representation after the first task (\textbf{org}).   Figure \ref{fig:t2} shows the projections from the different variants after learning the second task compared to the original projections. It can be seen that the Adaptive and Non Adaptive variants of our method try to preserve the projections from this subset. The adaptive projections are closer to the original one, if we look closely, while Finetuning projections starts drifting away from where they were. After the third task, as shown in figure \ref{fig:t3}, the Adaptive projections are closer to the original ones than the Non Adaptive that considers this subset as part of the full task being preserved and tries to prevent forgetting them as well. Finetuning started destroying the learned topology of this subset and lies further apart. However, when it comes to the fourth task, we see that it is a quite challenging and hard task. The forgetting appears more severe than before and preservation of the projections become even harder. Nevertheless,  the Adaptive {\tt MAS} and Non Adaptive {\tt MAS} still preserve the topology of the learned projections. The Adaptive projections lie closer and look more similar to the originals than the Non Adaptive MAS. Finetune, on the other hand, forgets completely about this subset and all the samples get projected in one point where it becomes quite hard to recognize their corresponding facts. 

\bibliographystyle{splncs04}
\bibliography{egbib}
\end{document}